\def\year{2022}\relax
\documentclass[letterpaper]{article} 
\usepackage{aaai22}  
\usepackage{times}  
\usepackage{helvet}  
\usepackage{courier}  
\usepackage[hyphens]{url}  
\usepackage{graphicx} 
\usepackage{natbib}  
\usepackage{caption} 
\DeclareCaptionStyle{ruled}{labelfont=normalfont,labelsep=colon,strut=off} 
\usepackage{algorithm}
\usepackage{algorithmic}

\usepackage{newfloat}
\usepackage{listings}
\floatstyle{ruled}
\newfloat{listing}{tb}{lst}{}
\floatname{listing}{Listing}
\usepackage{chngpage}
\usepackage[utf8]{inputenc} 
\usepackage{url}            
\usepackage{booktabs}       
\usepackage{nicefrac}       
\usepackage{microtype}      
\usepackage{subfig}
\usepackage{multirow}
\usepackage{amsmath}
\usepackage{amssymb}
\usepackage{amsthm}
\usepackage{amsfonts}
\usepackage{mathtools}
\usepackage{bbm}
\usepackage{xparse}
\usepackage[version=3]{mhchem}

\usepackage{xspace}

\usepackage{paralist}

\newcommand{\phii}{e}
\newcommand{\xii}{\boldsymbol{\omega}}

\newcommand{\bn}{\mathbf{n}}

\newcommand{\bmu}{\boldsymbol{\mu}}

\newcommand{\bx}{\mathbf{x}}

\newcommand{\by}{\mathbf{y}}

\newcommand{\bz}{\mathbf{z}}

\newcommand{\boldeta}{\boldsymbol{\eta}}
\newcommand{\kk}{t}
\newcommand{\la}{\mathbf{\ell}}

\newenvironment{prevproof}[1]{\noindent {\em {Proof of {#1}:}}}{\hfill $\square$\vskip \belowdisplayskip}

\newtheoremstyle{break}
  {}
  {}
  {\itshape}
  {}
  {\bfseries}
  {}
  {\newline}
  {}

\newtheorem{theorem}{Theorem}
\newtheorem{proposition}[theorem]{Proposition}

\newtheorem{lemma}[theorem]{Lemma}

\newcommand{\suppl}{supplementary material}

\DeclarePairedDelimiterX{\infdivx}[2]{(}{)}{%
  #1\;\delimsize\|\;#2%
}
\newcommand{\kld}[2]{\ensuremath{{\rm KL}\infdivx{#1}{#2}}\xspace}

\title{How Good are Low-Rank Approximations in Gaussian Process Regression?}
\author {
    Constantinos Daskalakis\textsuperscript{\rm 1}, Petros Dellaportas\textsuperscript{\rm 2, 4, 5}, Aristeidis Panos\textsuperscript{\rm 3}
}
\affiliations{
    \textsuperscript{\rm 1}CSAIL, Massachusetts Institute of Technology, USA\\
    \textsuperscript{\rm 2}University College London, UK\\
    \textsuperscript{\rm 3}University of Warwick, UK\\
    \textsuperscript{\rm 4}Athens University of Economics and Business, Greece\\
    \textsuperscript{\rm 5}The Alan Turing Institute, UK\\
    costis@csail.mit.edu, p.dellaportas@ucl.ac.uk, ares.panos@warwick.ac.uk
}

\begin{document}
\maketitle

\begin{abstract}
We provide  guarantees  for approximate Gaussian process regression resulting from two common low-rank kernel approximations: based on random Fourier features, and based on truncating the kernel's Mercer expansion.  In particular, we bound the Kullback–Leibler divergence between an exact Gaussian process and one resulting from one of the afore-described low-rank approximations to its 
kernel, as well as between their corresponding predictive densities. We provide experiments on both simulated data and standard benchmarks showing the effectiveness of our theoretical bounds. 

\end{abstract}


\section{Introduction}
\label{sec:introduction}

Gaussian processes (GPs) have long been studied in probability and statistics; see e.g.~\citet{williams2006gaussian}. In Bayesian inference, they provide a canonical way to define a probability distribution over functions, which can be used as a prior to build probabilistic frameworks for quantifying uncertainty in prediction. Among many applications, they have been a method of choice for hyperparameter tuning in deep learning~\cite{snoek2012practical}. 

In the simplest setting of GP regression, which is the focus of this paper, a measure over functions $f: \bx \mapsto y$ is defined such that, for any collection $X=(\bx_1,\ldots,\bx_N)$ of feature vectors,  their corresponding responses $\by=(y_1,\ldots,y_N)$ are jointly Gaussian, with zero mean and covariance matrix $K({k_\theta,X}):=(k_\theta(\bx_i,\bx_j))_{ij}$, where $k_\theta(\cdot,\cdot)$ is a positive semidefinite kernel indexed by some parameter vector $\theta$. A common inferential practice is to assume that we do not observe the Gaussian sample directly but additional  noise drawn from a zero-mean isotropic Gaussian is added to it prior to our observation.  Bayesian inference then proceeds by estimating~$\theta$ and the noise variance as well as computing predictive distributions of unobserved responses~$\by^*$ corresponding to a collection of new feature vectors~$X^*$ of interest.  These inference tasks require computing
the inverse and determinant of the covariance matrix $K({k_\theta,X})$, which naively costs $O(N^3)$ operations (or more precisely matrix multiplication time), making the inferential framework hard to scale computationally beyond a few thousand observations.

The computational burden 
of  GP inference has motivated a large body of work on faster, approximate  inference frameworks, as surveyed in~\citet{liu2020gaussian}. Many rely on the Nystr\"{o}m method, identifying for this purpose a set of ``inducing inputs'' on the input (i.e.~feature vector) domain~\citep{quinonero2005unifying,snelson2006sparse,titsias2009variational,williams2001using,hensman2013gaussian}, or the spectral domain   \citep{lazaro2010sparse,gal2015improving,hensman2017variational}. Other approaches are based on approximating the kernel by truncating its Mercer expansion~\citep{ferrari1999finite,solin2020hilbert}, or using random features \citep{cutajar2017random}. For more discussion see Section~\ref{sec:related}.

The motivating question for this work is that, while there is substantial work providing approximation guarantees for various low-rank kernel approximations with respect to different metrics, the impact of such approximation guarantees to the quality of approximate GP inference is not sufficiently understood. E.g.~many works provide
{\em entry-wise} approximation guarantees between a given kernel and an approximate one constructed via the Nystr\"om method, random features, Mercer expansion truncation, or other approximation technique; see e.g.~\citet{rahimi2008random,cortes2010impact,yang2012nystrom}. However, it is unclear how to translate such entry-wise guarantees to meaningful approximation guarantees relating GP inference using an exact kernel to inference using an approximate kernel. 

Recent work by~\citet{burt2019rates} pursued an investigation similar to ours for the sparse variational GP regression framework~\citep{titsias2009variational,hensman2015scalable}. They provide bounds for the Kullback–Leibler (KL) divergence between the true posterior distribution and one obtained using inducing inputs in the above framework. 

The goal of our work is to provide bounds for the impact to GP inference of two other prominent low-rank kernel approximation methods, based on random  features (Thm~\ref{thm:KL divergence instantiation random fourier features}) and on truncating the kernel's Mercer expansion (Thm~\ref{thm:KL divergence instantiation mercer expansion}). In particular, we provide bounds on the KL divergence between the marginal likelihood of an idealized GP with covariance matrix $K({k_\theta,X})$ and the marginal likelihood of a  GP with a low-rank covariance matrix~$\Sigma$ obtained from $k_\theta$ and $X$ using random features, or truncating $k_\theta$'s Mercer expansion. We quantify the KL divergence in terms of the rank of $\Sigma$. We show that moderate values of the rank suffice to bring the KL divergence below any desired threshold $\varepsilon N$, where $\varepsilon >0$. We obtain similar bounds for the KL divergence between the predictive densities of the exact and approximate GP, and we also bound the error between the predictive mean vectors and between the predictive covariance matrices computed using the exact vs using the approximate GP.  In the balance, our work provides theoretical grounding for the use of two common low-rank kernel approximations in GP regression, quantifying the inferential loss suffered in exchange for the computational benefit of  working with a low-rank kernel, as discussed in Sec~\ref{sec:constant rank Gaussian Process process}. 

In Sec~\ref{sec:rates}, we provide experiments investigating the effectiveness of our theoretical bounds in capturing the dependence of the KL approximation on the  dimension of the input features and the rank of $\Sigma$. In particular, by comparing the blue and green curves of Fig \ref{fig:theorem_3_5_ranks} we validate our theoretical results suggesting that the Gaussian kernels  require lower rank approximations to achieve a desired threshold $\varepsilon N$ when compared with Mat\'ern kernels with the same feature vector dimensions. Similarly, by comparing the solid and dotted curves of Fig \ref{fig:theorem_3_5_ranks} we validate our theoretical results suggesting that the Mercer approximations  require lower rank kernels to achieve a desired threshold $\varepsilon N$ when compared with a random feature approximations with the same feature vector dimensions.  Moreover, our theoretical bounds suggest that, for a fixed rank of $\Sigma$, approximating the Gaussian kernel using random features results in worst KL approximation compared to  approximating it by truncating its Mercer expansion. This is indeed reflected in our experiments on simulated data, when comparing the blue curves of the two panels of Fig~\ref{fig:theorem_3_5_instance}. Similarly, our theoretical bounds suggest that truncating the Mercer expansion of the Gaussian kernel provides better  approximation compared to truncating the expansion of the Mat\'ern kernel, and this is indeed reflected when comparing the blue and green curves of the right panel of Fig~\ref{fig:theorem_3_5_instance}.  

In a series of real data experiments in Sec \ref{sec:real data experiments}, we illustrate how low-rank approximations perform with different kernels and different ranks.  The results indicate that Mercer approximations outperform random Fourier features and they perform similarly with the sparse Gaussian process regression (SGPR) of \citet{titsias2009variational}, analyzed theoretically by~\citet{burt2019rates}. The better performance of Mercer compared to Fourier is consistent with our theoretical bounds. The similar performance of Mercer and SGPR is also consistent with  theory, as per our comparison to~\citet{burt2019rates} in Sec~\ref{sec:related}.

\paragraph{Paper Roadmap.} Sec~\ref{sec:related} discusses further  related work. Sec~\ref{sec:methodology} presents the basic GP regression setting, and well-known facts about the computational benefits of using low-rank kernel approximations. In Sec~\ref{sec:approximation guarantees}, we provide our theoretical results for the inferential impact to GP regression of using low-rank kernel approximations, based on random features (in Sec~\ref{sec:DFGP}) and based on truncating the Mercer expansion of the kernel (in Sec~\ref{sec:DMGP}). In both cases, we provide bounds on the KL divergence between a GP and one obtained by a low-rank approximation to its kernel. In Sec~\ref{sec:appx predictive} we state that these approximate guarantees are extended for the corresponding predictive densities. 
In Sec~\ref{sec:experiments}, we provide  experiments whose goal is two-fold: to illustrate our theoretical guarantees in simulated data scenarios and to investigate the practical performance of our studied kernel approximations, as suggested by our theoretical bounds, on a broad collection of standard benchmarks. 


\section{Related work}\label{sec:related}

The challenge of scaling up GP inference is  well-recognized and well-explored. We have already provided several references on approximate GP inference using inducing inputs and  kernel approximations. Theoretical guarantees for GP approximations with finite models, have been provided in \cite{zhu1997Gaussian,ferrari1999finite}, where the notion of Mercer truncation is utilized to provide similar results over the choice of the approximating rank $r$. In both cases, the quality of their approximation is expressed in terms of expected mean squared error. However, note that their bounds crucially depend on a ``large $N$'' assumption, as several sums are approximated by integrals in their development. Thus their bounds on mean squared error are only approximate and they do not quantify what is the loss resulting from their large $N$ assumption.

The approximation error resulting from low-rank approximations based on random Fourier features has been recently studied by~\citet{hoang2020revisiting}. In comparison to their results, our bound of Thm~\ref{thm:KL divergence instantiation random fourier features} is much more general as their guarantees require that the input feature vectors are sampled from a Gaussian mixture and also that the mixture components are (i) well-separated and (ii) they contribute exponentially decaying proportions of the points. In contrast, Thm~\ref{thm:KL divergence instantiation random fourier features} makes no distributional assumption about the input points.

Finally, a similar to ours theoretical investigation  has been pursued by~\citet{burt2019rates} for the different method of sparse variational GP regression of \citet{titsias2009variational,hensman2015scalable}. They provide bounds on the number of inducing inputs necessary to bring the KL divergence between the true GP posterior and the variational distribution obtained by the use of inducing inputs  below a desired threshold. For the Gaussian kernel, the required number of inducing inputs scales logarithmically in the number $N$ of training inputs, while for the Mat\'ern kernel it scales polynomially. While their paper and ours bound different quantities, our bounds from Thm~\ref{thm:KL divergence instantiation mercer expansion} are quantitatively similar to their bounds in Cor~22 for the Gaussian kernel, and  our bounds have a better dependence on $N$ compared to their bounds for the Mat\'ern kernel in Cor~25. (To compare set $\varepsilon = \gamma/N$ in our bounds or $\gamma = \varepsilon N$ in their bounds.)






\section{Preliminaries}
\label{sec:methodology}



\subsection{GP regression}\label{sec:basics}

In GP regression, we assume that response variables $\by =(y_i)_{i=1}^N \in \mathbb{R}^N$ corresponding to a collection of $D$-dimensional feature vectors $X = (\bx_i)_{i=1}^N \in \mathbb{R}^{N \times D}$ are noisy evaluations of some random function $f(\cdot)$,
i.e.~$y_i$ is a noisy observation of $f(\bx_i)$.
We take the noise, $y_i-f(\bx_i)$, for each data entry $i$ to be independent Gaussian with mean $0$ and variance~$\sigma^2$. Moreover, we place a GP prior over $f(\cdot)$, with zero mean and kernel $k_\theta(\cdot,\cdot)$, so that the collection of function values $f(X):=(f(\bx_i))_{i=1}^N$ has a joint Gaussian distribution with zero mean and covariance matrix $K(k_\theta,X):=(k_\theta(\bx_i,\bx_j))_{ij}$.  

The afore-described assumptions on the data generation process can be used in a regression setting in order to make predictions as follows. Suppose that we are given a collection of train feature vectors $X_{\rm train}\in \mathbb{R}^{N_{\rm train} \times D}$ and corresponding responses $\by_{\rm train}$ and our goal is to use this training data to predict the responses $\by_{\rm test}$ on a collection of test feature vectors $X_{\rm test}\in \mathbb{R}^{N_{\rm test} \times D}$. Setting  $A_{\rm train}=K(k_\theta,X_{\rm train}) + \sigma^2 I_{N_{\rm train}}$, the log-marginal likelihood of the training data becomes 
$\log p( \by_{\rm train} | X_{\rm train} ) = -\frac{1}{2} \by_{\rm train}^{\top} A_{\rm train}^{-1} \by_{\rm train} - \frac{1}{2} \log| A_{\rm train} | - \frac{N_{\rm train}}{2} \log(2\pi)$.  
Conditioning on the training data, the distribution of $\by_{\rm test}$ is normal with mean and variance given by the following:
\begin{itemize}
\item $\mathbb{E}(\by_{\rm test} | \by_{\rm train}) = K(k_\theta,X_{\rm test},X_{\rm train}) A_{\rm train}^{-1} \by $
\item $\mathbf{Var}(\by_{\rm test} | \by_{\rm train}) =K(k_\theta,X_{\rm test}) + \sigma^2 I_{N_{\rm test}}$\\ 
$~~~~~~~~~~~-K(k_\theta,X_{\rm test},X_{\rm train}) A_{\rm train}^{-1} K(k_\theta,X_{\rm test},X_{\rm train})^{\top}$
\end{itemize}
where $K({k_\theta, X_{\rm test},X_{\rm train}}):=(k_\theta(\bx_{{\rm test},i},\bx_{{\rm train},j}))_{ij}$.

\subsection{Low-rank GP regression}
\label{sec:constant rank Gaussian Process process}

Consider the setting of Sec~\ref{sec:basics}, and additionally suppose that the kernel function $k_{\theta}(\cdot,\cdot)$ is low-rank in the sense that there exists a feature map $\phi:\mathbb{R}^D \rightarrow \mathbb{R}^r$ such that for all $\bx,\bx' \in \mathbb{R}^D$: $k_{\theta}(\bx,\bx')=\langle \phi(\bx), \phi(\bx') \rangle,$ where $\langle \cdot,\cdot \rangle$ is the Euclidean inner product.  It follows that the kernel matrix $K(k_\theta,X)$ computed on a collection of feature vectors $X=(\bx_i)_{i=1}^N$ can be written as $K(k_\theta,X)=\Xi \Xi^{\top}$, where $\Xi$ is an $N \times r$ matrix whose rows are the vectors $\phi(\bx_{i})$, for $i=1,\ldots N$. As such, we get that the covariance matrix of the training data $\by_{\rm train}$  is $A_{\rm train}=\Xi_{\rm train} \Xi_{\rm train}^{\rm \top} + \sigma^2 I_{N_{\rm train}}$. We can then  use the Woodbury matrix inversion lemma and the Sylvester determinant theorem to obtain explicit forms for the inverse of $A_{\rm train}$ and its determinant:
    $A_{\rm train}^{-1} =  \sigma^{-2} I_{N_{\rm train}} - \sigma^{-2} \Xi_{\rm train} (\sigma^{2} I_r 
    + \Xi_{\rm train}^{\top} \Xi_{\rm train} )^{-1} \Xi_{\rm train}^{\top} $, and     $|A_{\rm train}| = \sigma^{2(N_{\rm train}-r)}|\sigma^2 I_r + \Xi_{\rm train}^{\top} \Xi_{\rm train}|$.
Since these identities involve inversion or determinant calculations of $r \times r $ matrices, by plugging them   into the expressions for the log-marginal likelihood of observations $\by$ and the mean and variance of the predictive density of future observations $\by^*$, we can, with the right ordering of operations, compute the log-likelihood and the predictive density in  $O(r^3+r^2N_{\rm train})$ time, i.e.~linear in $N_{\rm train}$, when $r$ is a constant.




\section{Approximation guarantees}
\label{sec:approximation guarantees}

We consider 
the setting of Sec~\ref{sec:basics}. In particular,  we suppose that $f(\cdot)$ is sampled from a GP with mean zero and kernel function $k_\theta:\mathbb{R}^D \times \mathbb{R}^D  \rightarrow \mathbb{R}$, 
and suppose that a collection $X=(\bx_i)_{i=1}^N$ of feature vectors maps to a collection of responses $\by=(y_i)_{i=1}^N$ sampled as follows
\begin{align}
\by \sim {\cal N}(0,K+\sigma^2 I_N), \label{eq:idealized model}
\end{align}
where $K:= K({k_\theta,X})\equiv (k_\theta(\bx_i,\bx_j))_{ij}$.

A well-studied topic in mathematics, statistics, and machine learning is approximating kernels with low-rank kernels.  Given a kernel function $k_\theta$, a long line of research has aimed to identify feature maps $\phi_{\theta,\varepsilon}:\mathbb{R}^D \rightarrow \mathbb{R}^r$ satisfying that, for a collection  of features vectors $X=(\bx_i)_{i=1}^N$,
\begin{align}
    K(k_\theta,X) \approx_{\varepsilon} \Sigma(\phi_{\theta,\varepsilon},X), \label{eq:low-rank kernel approximation guarantee}
\end{align}
where $\Sigma(\phi_{\theta,\varepsilon}, X)=(\phi_{\theta,\varepsilon}(\bx_i)^{\top} \phi_{\theta,\varepsilon}(\bx_j))_{ij}$. In~\eqref{eq:low-rank kernel approximation guarantee}, we have left the notion of approximation ``$\approx_{\varepsilon}$'' intentionally vague, for now, as there are many notions of approximation that have been pursued in the literature. We will soon discuss some instantiations. The parameter $\varepsilon$ is a tunable parameter controlling the quality of the approximation.

The goal of this paper is to quantify the loss of using the approximate kernel $\Sigma(\phi_{\theta,\varepsilon},X)$ in place of the original kernel $K(k_\theta,X)$  for the purposes of GP regression. In particular, we want to compare, in some precise sense, doing inference using the ``idealized model''~\eqref{eq:idealized model} versus an approximate model, which samples responses $\by=(y_i)_{i=1}^N$ for a collection of feature vectors  $X=(\bx_i)_{i=1}^N$ as follows:
\begin{align}
\by \sim {\cal N}(0,\Sigma+\sigma^2 I_N), \label{eq:approximate model}
\end{align}
where $\Sigma:=\Sigma(\phi_{\theta,\varepsilon},X)$. Notice that $\Sigma$ can be written as $\Sigma = \Xi \Xi^{\rm \top}$, where $\Xi$ is a $N \times r$ matrix whose rows are $\phi_{\theta,\varepsilon}(\bx_i)$ for $i=1,\ldots,N$. Thus, $\Sigma$ is a rank-$r$ matrix and, as discussed in Sec~\ref{sec:constant rank Gaussian Process process}, GP regression under~\eqref{eq:approximate model} is computationally cheap when $r$ is small. Our goal is to  quantify the inferential loss suffered in exchange for the computational benefit of  working with a low-rank kernel. 

The sense in which we aim to quantify the inferential loss is by bounding the KL divergence between the marginal likelihood under~\eqref{eq:approximate model} and under~\eqref{eq:idealized model}. In Secs~\ref{sec:DFGP} and~\ref{sec:DMGP}, we provide such bounds for two common low-rank kernel approximation methods, based on random features and Mercer expansion truncation respectively. In Sec~\ref{sec:appx predictive}, we show that our bounds also bound the KL divergence between the predictive densities, as well as the prediction error.


We start with a generic result, providing bounds on the KL divergence between Gaussians whose covariance matrices have special structure. 


\begin{proposition}[Proof in the \suppl] \label{thm:KL approximation theorem}
Suppose that $\Sigma_1$ and $\Sigma_2$ are $N \times N$ positive definite (symmetric) matrices, such that $(1+\gamma)\Sigma_1-\Sigma_2$ is positive semi-definite for some $\gamma \ge 0$. Then
\begin{align}
    & \kld{\mathcal{N}(0,\Sigma_1)}{\mathcal{N}(0,\Sigma_2)} \le \nonumber \\ 
    & \frac{1}{2}{\rm Tr}(\Sigma_2^{-1/2}(\Sigma_1-(1-\gamma)\Sigma_2)\Sigma_2^{-1/2}).~~~~~ \label{eq:costas bound 1}
\end{align}

If additionally $\Sigma_2 \succeq (1+\gamma)^{-1} \Sigma_1$, then we obtain
\begin{align}
\kld{\mathcal{N}(0,\Sigma_1)}{\mathcal{N}(0,\Sigma_2)} &\le {\gamma N}.~~~~~~~~~~~~~~~~~~~ \label{eq:costas bound 2}
\end{align}

If $\Sigma_1=\sigma^2 I_N + K_1$ and $\Sigma_2=\sigma^2I_N+K_2$, where $K_1$ and $K_2$ are positive semi-definite, $\sigma^2>0$, and $(1+\gamma)\Sigma_1-\Sigma_2$ is positive semi-definite, then
\begin{align}
    & \kld{\mathcal{N}(0,\Sigma_1)}{\mathcal{N}(0,\Sigma_2)} \le \nonumber\\
    & \frac{1}{2 \sigma^2} {\rm Tr}(K_1-(1-\gamma)K_2+\gamma \sigma^2 I_N). \label{eq:costas bound}
\end{align}
\end{proposition}

 In the next sections we will instantiate Prop~\ref{thm:KL approximation theorem} by taking $K_1=K(k_\theta,X)$ and $K_2=\Sigma(\phi_{\theta,\varepsilon},X)$, resulting respectively in the idealized data generation process of~\eqref{eq:idealized model} and the approximate one of~\eqref{eq:approximate model}. Our theorem states that the KL divergence between these two processes is controlled by~\eqref{eq:costas bound 1}--\eqref{eq:costas bound}, which as we will see   next  can become smaller than any desired~$\varepsilon N$ for relatively modest values of the rank $r$, namely poly-logarithmic in $N$ (Thm~\ref{thm:KL divergence instantiation random fourier features}), or even an absolute constant (Thm~\ref{thm:KL divergence instantiation mercer expansion}),  whenever the dimension~$D$ is an absolute constant.

\subsection{Guarantees for random feature approximation} \label{sec:DFGP}

A common framework for low-rank kernel approximations defines a parametric family of functions $e_{\boldeta}:\mathbb{R}^D \rightarrow \mathbb{R}$ and a distribution $p(\boldeta)$ over $\boldeta$, picking a random feature map $\phi(\bx)=(e_{\boldeta_1}(\bx),\ldots,e_{\boldeta_r}(\bx))$ by sampling $\boldeta_1,\ldots,\boldeta_r \sim p(\boldeta)$. The goal is that the resulting feature map $\phi$ results in a good approximation of some target kernel matrix $K(k,X)$ by $\Sigma(\phi,X)$, as discussed earlier in this section.

For example, the celebrated work by~\cite{rahimi2008random} exploits Bochner's theorem for shift invariant kernels to derive from it  a kernel-specific density $p(\boldeta)$ that it uses in conjunction with the family of cosine functions $e_{\boldeta}(\cdot)$ with frequency and phase determined by $\boldeta$. Specifically $e_{\boldeta}(\cdot)$ is derived from  a random Fourier feature with spectral frequency $\boldeta$; see also \cite{cutajar2017random}. 

However, the kernel approximation guarantees obtained by~\cite{rahimi2008random} (as well as by much work in this literature) only bound the {\em element-wise} distance between the kernel matrices $K(k,X)$ and $\Sigma(\phi,X)$. To bound the KL divergence between~\eqref{eq:idealized model} and~\eqref{eq:approximate model} such entry-wise bounds are insufficient. Rather, we need a spectral approximation of $K(k,X) + \sigma^2I$ by $\Sigma(\phi,X) + \sigma^2 I$, as per Prop~\ref{thm:KL approximation theorem}. Making use of  spectral approximations by~\cite{avron2017random} for modified Fourier features, we show that  the KL divergence between~\eqref{eq:idealized model} and~\eqref{eq:approximate model}  can indeed be controlled for the Gaussian kernel. We provide our statement for the Gaussian kernel with the same fixed scaling in every direction for notational simplicity. It extends to the general Gaussian kernel with different scaling per direction in an obvious way (namely by rescaling coordinates).

\begin{theorem}[Proof in the \suppl] \label{thm:KL divergence instantiation random fourier features}

Consider the $D$-dimensional Gaussian kernel $k(\bx,\bx')=\exp(-2\pi^2||\bx-\bx'||_2^2)$, and the kernel matrix $K=K(k,X)=(k(\bx_i,\bx_j))_{ij}$, where $X=(\bx_1,\ldots,\bx_N)$ is a collection of points in $\mathbb{R}^D$ such that, for some $R>0$, $||\bx_i-\bx_j||_{ \infty }\le R, \forall i,j$. Suppose $D \le 5 \log(N/\sigma^2)+1$ and $\varepsilon \in (0,0.5)$. There exists (a samplable in $O(D)$-time) distribution $p(\boldeta)$ and a parameterized family $e_{\boldeta}(\cdot)$ of {\em modified} Fourier Features such that, if we take $r \ge \Omega\left({R^D \over\varepsilon^2}  (\log{N\over \sigma^2})^{2D} \log({N\over \delta})\right)$ random $\boldeta_1,\ldots,\boldeta_r \sim p(\boldeta)$ and define the rank-$r$ matrix $\Sigma=\Sigma(\phi,X)$ using the feature map $\phi(\bx)=(e_{\boldeta_1}(\bx),\ldots,e_{\boldeta_r}(\bx))$, then with probability at least $1-\delta$, the KL divergence from distribution~\eqref{eq:approximate model} to distribution~\eqref{eq:idealized model}  is at most $\varepsilon N$.
\end{theorem}




\subsection{Guarantees for Mercer truncation approximation} \label{sec:DMGP}

In this section, we discuss an alternative approach for obtaining low-rank kernel approximations, based on
truncating the Mercer expansion of the kernel~\citep{mercer1909functions}, and our associated approximation guarantees when this low-rank kernel approximation  is used in GP regression. 

Suppose that $k_\theta$ is a Mercer kernel on some probability space ${\cal X} \subseteq \mathbb{R}^D$ with probability measure $\mu$, which means that $k_\theta(\cdot,\cdot)$ can be written as:
\begin{equation}
k_{\theta}(\bx,\bx') = \sum_{\kk=1}^{\infty} \lambda_{\kk} \phii_{\kk}(\bx) \phii_{\kk}(\bx'), \label{eq:mercer expansion}
\end{equation}
where $(\lambda_\kk)_{\kk \in \mathbb{N}}$ is a sequence of summable non-negative, non-increasing numbers, i.e.~\emph{eigenvalues}, and $(\phii_\kk)_{\kk \in \mathbb{N}}$ is a family of mutually orthogonal unit-norm functions with respect to the inner product $\langle f, g \rangle = \int_{\cal X}f(\bx)g(\bx)d\mu(\bx)$, defined by $\mu$, i.e.~\emph{eigenfunctions}. Now suppose that $X=(\bx_i)_{i=1}^N$ is a collection of vectors $\bx_i \in {\cal X}$. It follows from Eq.~\eqref{eq:mercer expansion} that the kernel matrix $K(k_{\theta},X)$ can be written as:
\begin{align}
    K(k_{\theta},X) \equiv \sum_{\kk=1}^{\infty} \lambda_{\kk} \xii_{\kk} \xii_{\kk}^{\top}, \label{eq:expansion of kernel matrix}
\end{align} 
where $\xii_{\kk}= (\phii_{\kk}(\bx_1),\phii_{\kk}(\bx_2),\ldots,\phii_{\kk}(\bx_N))$, for all $\kk \in \mathbb{N}$.
Recall that the sequence $(\lambda_\kk)_{\kk}$ is summable so $\lambda_\kk \rightarrow 0$ as $\kk \rightarrow \infty$. The rate of decay is very fast for many kernels. For example, the decay is exponentially fast for the Gaussian kernel, and polynomially fast for the Mat\'ern kernel when the input distribution is compact or concentrated. These are standard facts (see e.g.~\cite{williams2006gaussian}), but for completeness we illustrate how to derive the eigendecomposition of the high-dimensional Gaussian kernel under a Gaussian input density in Sec~\ref{sec:gaussian kernel mercer illustration} of the supplement.

The fast decay of the eigenvalues motivates approximating $K(k_{\theta},X)$ by keeping the first few terms of~\eqref{eq:expansion of kernel matrix}. In our theorem, we quantify the  impact of that truncation to GP regression in terms of the KL divergence between the data likelihood of the GP process with kernel $K(k_{\theta},X)$ and the GP process with the truncated kernel.

\begin{theorem}[Proof in the \suppl] \label{thm:KL divergence instantiation mercer expansion}
Let $k(\cdot,\cdot)$ be a Mercer kernel on probability space $({\cal X},\mu)$ with $k(\bx,\bx)\le B$, for all $\bx \in {\cal X}$. Let $X=(\bx_1,\ldots,\bx_N)$ comprise samples from $\mu$, let $K=K(k,X)$ (which satisfies~\eqref{eq:expansion of kernel matrix}), and let $\Sigma=\sum_{\kk=1}^{r} \lambda_{\kk} \xii_{\kk} \xii_{\kk}^{\top}$, for some $r \in \mathbb{N}$. (So the rank of $\Sigma$ is $r$.) With probability at least $1-\delta$ (with respect to the samples $X$),
the KL divergence from distribution~\eqref{eq:approximate model} to distribution~\eqref{eq:idealized model} is at most
\begin{align}
{N \over 2\sigma^2} \cdot \left( \Lambda_{>r} + \sqrt{{B\Lambda_{>r} \over N \delta} }\right), \label{eq:KL bound from mercer truncation}
\end{align}
where $\Lambda_{>r}=\sum_{\kk>r}\lambda_\kk$. Two example instantiations of the bound are as follows:
\begin{itemize}
    \item Suppose $k(\bx,\bx')=\exp(-2\pi^2||\bx-\bx'||_2^2)$ is the multi-variate Gaussian kernel over $\mathbb{R}^D$, endowed with a Gaussian density $\mu(\bx) = ({2  \pi / {R^2}})^{D \over 2} \exp(-2 \pi^2 ||\bx||_2^2/{R^2})$, where $R>0$. For any  absolute constant $0<c<1$, choosing rank 
    \begin{align*}
        r = \begin{cases}
        &\hspace{-13pt}\left(\Omega(RD\log(RD \vee e) + R \log{1 \over \varepsilon \sigma \delta})\right)^{D},~\text{if $R\ge c$}\\
        &\hspace{-13pt}\Biggl( \Omega \Biggl( {D \over \log{1 \over R}} \Bigl( \log{D \over \log{1 \over R}} \vee \log{2 \over R^2} \Bigr) + {\log{1 \over \varepsilon \sigma \delta} \over \log{1 \over R}} \Biggr)\Biggr)^D \hspace{-8pt},~\text{ow}
        \end{cases}
    \end{align*}

    makes~\eqref{eq:KL bound from mercer truncation} at most $\varepsilon N$. 
    In both bounds the constant hidden by the $\Omega(\cdot)$ notation depends on $c$ and no other parameter. Moreover, the bound easily extends  to Gaussian kernels with different length scales per dimension and other product Gaussian input measures $\mu$.
    
    \item Suppose $k(\bx,\bx')$ is a  Mat\'ern kernel with parameter $\nu>0$ and length scale $\alpha>0$ over $\mathbb{R}^D$,\footnote{Specifically the kernel takes the form $$k(\bx,\bx')={2^{1-\nu} \over \Gamma(\nu)}\left({\|\bx-\bx'\|_2 \over \alpha}\right)^{\nu} K_{\nu}\left({\|\bx-\bx'\|_2 \over \alpha}\right),$$
    where $K_{\nu}$ is a modified Bessel function and $\alpha = \ell / \sqrt{2 \nu}, \ell>0$; see Chapter 4 of~\citet{williams2006gaussian}.} endowed with a bounded measure $\mu(\bx)$ over a bounded set. Then choosing rank $r \ge A \left({1 \over \varepsilon \sigma \delta} \right)^{\Omega(D/\nu)}$ makes~\eqref{eq:KL bound from mercer truncation} at most $\varepsilon N$, for some constant $A$ that depends on $\nu, \alpha, D$ and the bounds on $\mu$ and its support, but does not depend on $\varepsilon,\sigma,\delta,N$.
\end{itemize}
\end{theorem}



\subsection{Approximation guarantees for GP regression} 
\label{sec:appx predictive}

There are two ways to use Thms~\ref{thm:KL divergence instantiation random fourier features} and~\ref{thm:KL divergence instantiation mercer expansion} to obtain bounds on the approximation error resulting from using an approximate GP model based on either random Fourier features or  truncating the Mercer expansion of the kernel.  Indeed, if we apply those theorems using $X=X_{\rm train}$ and $\by=\by_{\rm train}$ we immediately get bounds on $\kld{P(\by_{\rm train})}{Q(\by_{\rm train})}$ where $P(\by_{\rm train})$ and $Q(\by_{\rm train})$ are respectively the densities of the training data under the exact GP and the approximate one. Indeed, in the settings of Thms~\ref{thm:KL divergence instantiation random fourier features} and~\ref{thm:KL divergence instantiation mercer expansion} and the choice of rank made in these theorems for $N=N_{\rm train}$, we get that with probability at least $1-\delta$ (with respect to the randomness in the sampling of modified Fourier Features in the setting of Thm~\ref{thm:KL divergence instantiation random fourier features} and the sampling of $X_{\rm train}$ in the setting of Thm~\ref{thm:KL divergence instantiation mercer expansion}):
$$0\le\kld{P(\by_{\rm train})}{Q(\by_{\rm train})} \le \varepsilon N_{\rm train}.$$
By the definition of KL divergence, this bound can be interpreted as a bound on the difference of the marginal likelihoods under the true and the approximate GP, in expectation over data sampled from the true GP. Indeed, we equivalently get that, with probability $\ge 1-\delta$:   $$0\le \mathbb{E}_{\by_{\rm train} \sim P}\left[\log P(\by_{\rm train})-\log Q(\by_{\rm train})\right]\le \varepsilon N_{\rm train}.$$


Moreover, it is straightforward to use Thms~\ref{thm:KL divergence instantiation random fourier features} and~\ref{thm:KL divergence instantiation mercer expansion} to obtain bounds on the Kullback–Leibler divergence between the {\em predictive densities} of unobserved responses $\by_{\rm test}$ on new features $X_{\rm test}$ corresponding to the exact GP and that obtained by either random Fourier features or by truncating the Mercer expansion of the kernel.  
Indeed, if we apply those theorems using $X = \begin{bmatrix} X_{\rm train} \\ X_{\rm test} \end{bmatrix} $ and $\by= \begin{bmatrix} \by_{\rm train} \\ \by_{\rm test} \end{bmatrix}$, we immediately get bounds on $\kld{P(\by_{\rm train} , \by_{\rm test})}{Q(\by_{\rm train} , \by_{\rm test})}$ where $P(\by_{\rm train} , \by_{\rm test})$ and $Q(\by_{\rm train} , \by_{\rm test})$ are the joint densities of the combined vector of observed and unobserved responses, under respectively the exact GP and the approximate GP. Indeed, in the settings of Thms~\ref{thm:KL divergence instantiation random fourier features} and~\ref{thm:KL divergence instantiation mercer expansion} and the choice of rank made in these theorems for $N = N_{\rm train} + N_{\rm test}=N_{\rm total}$, we get that with probability at least $1-\delta$ (w.r.t.~the randomness in the sampling of modified Fourier Features in the setting of Thm~\ref{thm:KL divergence instantiation random fourier features} and the sampling of $X_{\rm train}$ in the setting of Thm~\ref{thm:KL divergence instantiation mercer expansion}):
\begin{align}\kld{P(\by_{\rm train} , \by_{\rm test})}{Q(\by_{\rm train} , \by_{\rm test})} \le \varepsilon N_{\rm total}. \label{eq:costas kourasi 50}
\end{align}
By the chain rule of KL divergence, the LHS of~\eqref{eq:costas kourasi 50} equals 
\begin{align*}
&~~~~~~~~~~\kld{P(\by_{\rm train})}{Q(\by_{\rm train})} \\
             &~~~~~~~~~~~~ + \kld{P(\by_{\rm test}|\by_{\rm train})}{Q(\by_{\rm test}|\by_{\rm train})}.
\end{align*}
Because of the non-negativity of Kullback–Leibler divergence, Eq~\eqref{eq:costas kourasi 50} implies that with probability at least $1-\delta$:
$$\kld{P(\by_{\rm test}|\by_{\rm train})}{Q(\by_{\rm test}|\by_{\rm train})}\le \varepsilon N_{\rm total}.$$
By the definition of conditional KL divergence, we equivalently get that with probability $\ge 1-\delta$ (w.r.t.~the randomness in the sampling of modified Fourier Features in the setting of Thm~\ref{thm:KL divergence instantiation random fourier features} and the sampling of $X_{\rm train}$ in the setting of Thm~\ref{thm:KL divergence instantiation mercer expansion}) the expected (w.r.t.~$\by_{\rm train} \sim P$) KL divergence between the predictive densities of the true and the approximate GPs are close:
\begin{align}
\mathbb{E}_{\by_{\rm train} \sim P}\left[\kld{P(\by_{\rm test}|\by_{\rm train})}{Q(\by_{\rm test}|\by_{\rm train})}\right]\le \varepsilon N_{\rm total}. \label{eq:costas kourasi 56}
\end{align}
In turn, by using Markov's inequality, we get that for $\eta$ of our choosing, with probability at least $1-\delta -\eta$ (w.r.t.~the sampling of both $\by_{\rm train}\sim P$ and the randomness in the sampling of modified Fourier Features in the setting of Thm~\ref{thm:KL divergence instantiation random fourier features} and the sampling of $X_{\rm train}$ in the setting of Thm~\ref{thm:KL divergence instantiation mercer expansion}):
\begin{align}\kld{P(\by_{\rm test}|\by_{\rm train})}{Q(\by_{\rm test}|\by_{\rm train})}\le {\varepsilon\over\eta} N_{\rm total}. \label{eq:costas kourasi 57}
\end{align}
Finally, since both $P(\by_{\rm test}|\by_{\rm train})$ and $Q(\by_{\rm test}|\by_{\rm train})$ are Gaussian distributions, our bounds from \eqref{eq:costas kourasi 56} and~\eqref{eq:costas kourasi 57} directly bound the error between the predictive mean vectors and between the predictive covariance matrices computed using the approximate vs using the true GP, as per the following proposition:
\begin{proposition}\label{prop:from KL to parameter bounds}
Consider arbitrary $N$-dimensional Gaussians ${\cal N}(\bmu_1,\Sigma_1)$ and ${\cal N}(\bmu_2,\Sigma_2)$. Suppose that $\Sigma_1$ and $\Sigma_2$ are non-singular, and suppose that $\kld{{\cal N}(\bmu_1,\Sigma_1)}{{\cal N}(\bmu_2,\Sigma_2)} \le \gamma$ for some $\gamma \ge 0$. Then
\begin{align}
    &{1 \over 2}(\bmu_2-\bmu_1)^{\rm T}\Sigma_2^{-1}(\bmu_2-\bmu_1) \le \gamma,\label{eq:costas mean bounds}\\
    &b(2\gamma)\cdot \Sigma_2 \preceq \Sigma_1 \preceq t(2\gamma) \cdot \Sigma_2,\label{eq:costas cov bounds}
\end{align}
where $b(2\gamma)$ and $t(2\gamma)$ are respectively the smallest and largest roots of $x-1-\ln(x)=2\gamma$. In particular, the Mahalanobis distance of $\bmu_1$ from $(\bmu_2,\Sigma_2)$ is bounded by $\sqrt{2\gamma}$ and $\Sigma_1$ and $\Sigma_2$ are spectrally close.\footnote{It can be shown that $b(2\gamma) \ge \max(1-2\sqrt{\gamma}, \exp(-1-2\gamma))$ and $t(2\gamma)\le 1+\max(\sqrt{8\gamma},8\gamma)$ so these explicit expressions can be plugged in place of $b(2\gamma)$ and $t(2\gamma)$ respectively in~\eqref{eq:costas cov bounds}.}
\end{proposition}
We set $P(y_{\rm test}|\by_{\rm train})$ in place of ${\cal N}(\bmu_1,\Sigma_1)$ and $Q(y_{\rm test}|\by_{\rm train})$ in place of ${\cal N}(\bmu_2,\Sigma_2)$ in Prop~\ref{prop:from KL to parameter bounds} and combine it with the bound of~\eqref{eq:costas kourasi 57} to get the following: 

\begin{itemize}
  \item With probability $\ge 1-\delta-\eta$ (w.r.t.~the randomness in the sampling of modified Fourier features in the setting of Thm~\ref{thm:KL divergence instantiation random fourier features} and the sampling of $X_{\rm train}$ in the setting of Thm~\ref{thm:KL divergence instantiation mercer expansion} as well as the sampling of $\by_{\rm train} \sim P$) the Mahalanobis distance between the predictive mean vector under the true and the approximate GP is at most $\sqrt{2\varepsilon/\eta N_{\rm total}}$. Moreover, the predictive covariances of the true and the approximate GPs are sandwiched as follows $b(2\varepsilon/\eta N_{\rm total})\cdot \Sigma_2 \preceq \Sigma_1 \preceq t(2\varepsilon/\eta N_{\rm total}) \cdot \Sigma_2$.
\end{itemize}
For the above bounds to be most effective, it makes sense to choose $\varepsilon$ to scale with $N_{\rm total}$, perhaps as $1/N_{\rm total}^{\kappa}$ for some $\kappa$. We note that depending on the choice of $\kappa$ the rank bound of Thm~\ref{thm:KL divergence instantiation random fourier features} may or may not be effective. (It is effective if $\kappa<1/2$). On the other hand, the rank bound of Thm~\ref{thm:KL divergence instantiation mercer expansion} remains effective regardless the choice of $\kappa$ as the appearance of $\varepsilon$ in the rank bound is milder. 

\section{Experiments}\label{sec:experiments}
We perform a series of simulated and real-data experiments for studying our theoretical bounds in practise. The low-rank approximation techniques of Secs \ref{sec:DFGP} and \ref{sec:DMGP} are refered to as  \emph{Fourier GP (FGP)} and \emph{Mercer GP (MGP)} respectively.

Inference for FGP is based on guidelines given by \cite{rahimi2008random}. We sample, for  even $r$, $r \over 2$ spectral frequencies $\boldeta_1,\ldots,\boldeta_{r \over 2}$ from the spectral density $p(\boldeta)$ of the 
 kernel, and compute the feature map $\phi(\bx):\mathbb{R}^D \rightarrow \mathbb{R}^{r}$, defined by
$\sqrt{\frac{2}{r}}[\cos(\boldeta_1^{\top} \bx),\ldots, \cos(\boldeta_{r \over 2}^{\top} \bx), \sin(\boldeta_1^{\top} \bx),\ldots, \sin(\boldeta_{r \over 2}^{\top} \bx) ]^{\top}$. 
The spectral frequencies are only sampled once, before training, and are  kept fixed throughout the optimization of the log-marginal likelihood. The spectral density of the Gaussian kernel~\eqref{eq:ard_kernel}, which  we use in our experiments, is 
$ p(\boldeta) =  \sqrt{| 2 \pi \Delta^{-1} |}\sigma_f^{-2}  \exp( - 2 \pi^2 \boldeta^{\top} \Delta^{-1} \boldeta )$.

Inference for MGP is straightforward when $D=1$. However, when $D>1$ the eigenvalues and eigenvectors are constructed as tensor products so their computational complexity is prohibitively large for even small values of $D$; see Sec~\ref{sec:gaussian kernel mercer illustration} of the supplement. Therefore, for our real data experiments we devise a simple computational trick to circumvent this problem as follows. We linearly project the $D$-dimensional features to a lower $d$-dimensional space via a weight matrix $W$ of dimension $D \times d$ that is readily estimated by maximizing the marginal log-likelihood. The resulting projected feature matrix $Z=(\bz_1,\bz_2,\ldots,\bz_N)$ has dimension $N \times d$.  Next,  we compute a low-rank  $\Sigma$ by keeping, from the Mercer expansion of the kernel, the top $r$, under some ordering, tensor products of the  eigenfunctions:
$\Sigma = \sum_{n=1}^r \lambda_n \xi_{n} \xi_{n}^{\top}$,   
where $\xi_{n}= [ \phii_{n}(\bz_1), \ldots, \phii_{n}(\bz_N) ]^{\top} \in \mathbb{R}^N$ and $e_{r}$ is the eigenfunction of the kernel indexed by $r$. The choice of the ordering of   
the tensor product of eigenvectors is first based on the total degree of the corresponding orders in each dimension and second on lexicographical order. 
Finally, note that the parameters $a_j$ in \eqref{input} of Sec~\ref{sec:gaussian kernel mercer illustration} of the supplement have to be prefixed or learnt from the data. We choose to keep them fixed with their values  set to $1 / \sqrt{2}$, which corresponds to a standard $d$-dimensional Gaussian measure over $\bz$. Thus, we also standardize the projected features $\bz$. 

\begin{figure}[h]
\centering
\includegraphics[width=8.5cm,height=4.5cm]{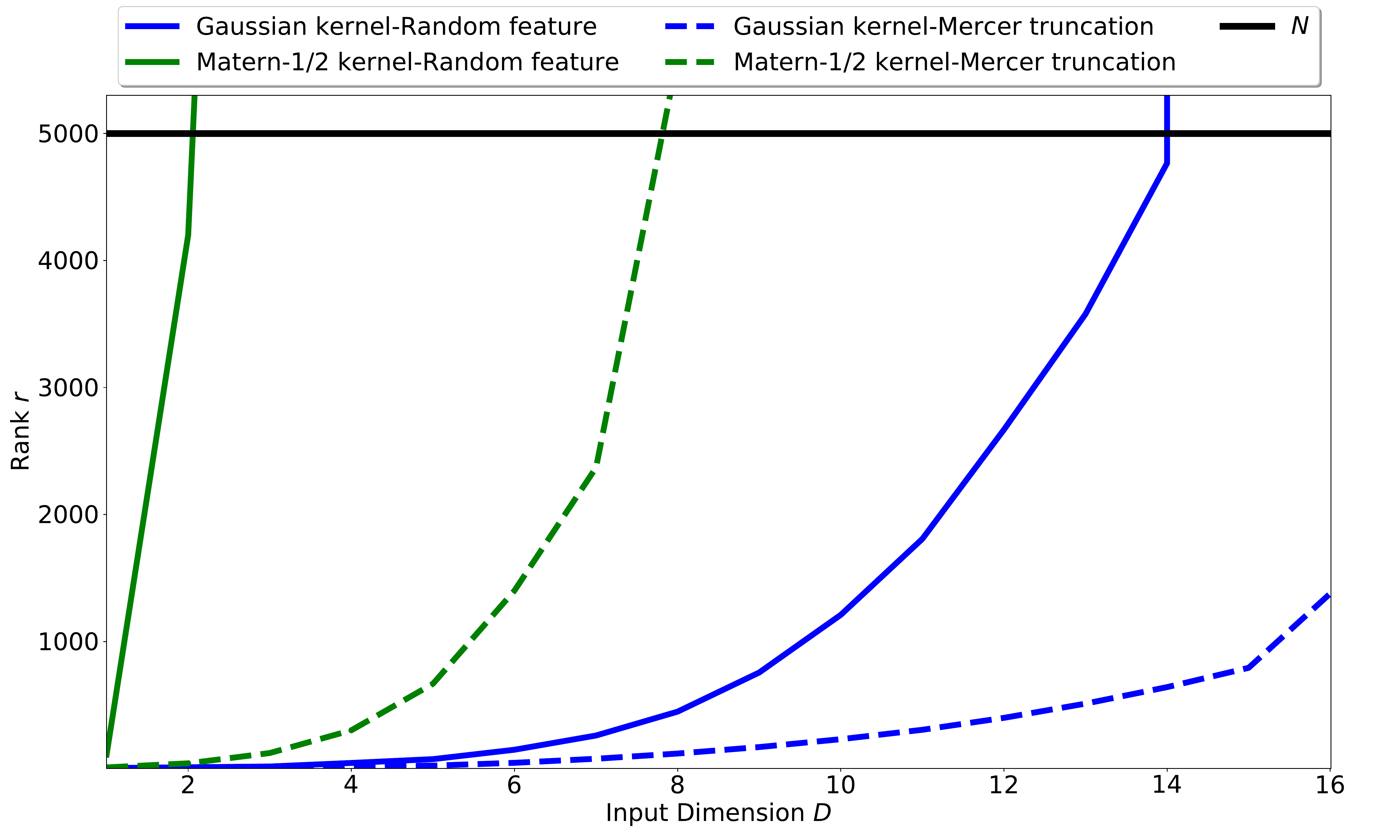} 
\caption{The lowest rank $r$ needed for $\Sigma$ such that $\kld{\mathcal{N}(0, K + \sigma^2 I_N)}{\mathcal{N}(0,\Sigma + \sigma^2 I_N)} \leq \varepsilon N$ as the input dimension $D$ increases, in the low-rank approximation settings of 
Thms~\ref{thm:KL divergence instantiation random fourier features} (solid lines) and~\ref{thm:KL divergence instantiation mercer expansion} (dashed lines).  We sample $N=5000$ points from $ \mathcal{N}(0,\frac{1}{16^2} I_D)$ for the GP  with the Gaussian kernel, and sample  from a $D$-dimensional $\mathcal{U}(-\frac{\sqrt{3}}{16}, \frac{\sqrt{3}}{16})$ for the  Mat\'ern-1/2 kernel. The values of $\sigma^2$ and $\varepsilon$ are set equal to $1$ and $10^{-2}$, respectively.
 For reference, we show the full rank line in black.} 
\label{fig:theorem_3_5_ranks}
\end{figure}



\begin{table*}[h!]
\begingroup
\setlength{\tabcolsep}{9.1pt} 
\begin{small}
\begin{sc}
\begin{tabular}{l c c c c c c c}
\toprule
&  \multicolumn{6}{c}{\textbf{Negative Log-Predictive Density}}  \\
\cmidrule(lr){2-8}
   &   Bike & Elevators   &   Super & Protein   &   Sarcos   &   KeggDir & 3DRoad      \\
$N_{\rm train}$ &   $ 15641$   &   $14939 $   &   $ 19136 $   &   $41157$   &   $44039$   &   $48071 $   &   $391386$   \\
 $N_{\rm test}$ &  $1738$ & $1660 $   &   $2127 $   &   $ 4573 $   &   $ 4894 $   &   $5342 $   &   $43488 $   \\
 $D$ &  $57$ & $ 18 $   &   $81$   &   $ 9 $   &   $ 21 $   &   $19 $   &   $3 $    \\
\midrule
SGPR6   &   $1.43 (0.01)$   &   $0.52 (0.01)$   &   $0.69 (0.00)$   &   $1.22 (0.00)$   &   $0.35 (0.01)$   &   $0.65 (0.01)$   &   $1.32 (0.01)$   \\
FGP6   &   $0.97 (0.02)$   &   $1.13 (0.18)$   &   $0.76 (0.01)$   &   $1.29 (0.01)$   &   $0.41 (0.00)$   &   $1.27 (0.01)$   &   $1.28 (0.00)$   \\
MGP6   &   $0.99 (0.01)$   &   $0.56 (0.01)$   &   $0.97 (0.01)$   &   $1.25 (0.00)$   &   $0.33 (0.00)$   &   $0.98 (0.00)$   &   $1.31 (0.01)$   \\
\midrule
SGPR10   &   $0.90 (0.28)$   &   $0.50 (0.01)$   &   $0.64 (0.01)$   &   $1.19 (0.00)$   &   $0.16 (0.02)$   &   $0.58 (0.01)$   &   $1.06 (0.00)$   \\
FGP10   &   $0.92 (0.01)$   &   $1.16 (0.24)$   &   $0.76 (0.00)$   &   $1.22 (0.01)$   &   $0.34 (0.00)$   &   $1.06 (0.01)$   &   $1.21 (0.00)$   \\
MGP10   &   $0.73 (0.01)$   &   $0.48 (0.01)$   &   $0.68 (0.01)$   &   $1.19 (0.00)$   &   $-0.04 (0.00)$   &   $0.67 (0.00)$   &   $1.15 (0.01)$   \\
\midrule
SGPR50   &   $0.19 (0.01)$   &   $0.46 (0.01)$   &   $0.57 (0.01)$   &   $1.11 (0.01)$   &   $-0.21 (0.00)$   &   $0.35 (0.01)$   &   $0.92 (0.00)$   \\
FGP50   &   $0.92 (0.01)$   &   $0.56 (0.01)$   &   $0.69 (0.01)$   &   $1.19 (0.00)$   &   $-0.11 (0.00)$   &   $0.71 (0.01)$   &   $1.06 (0.00)$   \\
MGP50   &   $0.29 (0.01)$   &   $0.45 (0.01)$   &   $0.51 (0.01)$   &   $1.15 (0.00)$   &   $-0.26 (0.00)$   &   $0.46 (0.00)$   &   $1.02 (0.00)$   \\
\midrule
SGPR100   &   $0.05 (0.01)$   &   $0.44 (0.01)$   &   $0.53 (0.01)$   &   $1.07 (0.01)$   &   $-0.30 (0.00)$   &   $0.27 (0.00)$   &   $0.86 (0.01)$   \\
FGP100   &   $0.87 (0.04)$   &   $0.53 (0.01)$   &   $0.64 (0.00)$   &   $1.16 (0.00)$   &   $-0.20 (0.00)$   &   $0.60 (0.01)$   &   $1.01 (0.00)$   \\
MGP100   &   $0.04 (0.01)$   &   $0.43 (0.01)$   &   $0.51 (0.01)$   &   $1.14 (0.00)$   &   $-0.29 (0.00)$   &   $0.39 (0.00)$   &   $0.85 (0.00)$   \\
\midrule
SGPR200   &   $0.02 (0.01)$   &   $0.43 (0.01)$   &   $0.50 (0.01)$   &   $1.01 (0.01)$   &   $-0.39 (0.00)$   &   $0.20 (0.01)$   &   $0.83 (0.01)$   \\
FGP200   &   $0.81 (0.03)$   &   $0.47 (0.01)$   &   $0.58 (0.01)$   &   $1.13 (0.01)$   &   $-0.30 (0.00)$   &   $0.47 (0.01)$   &   $0.89 (0.00)$   \\
MGP200   &   $0.05 (0.01)$   &   $0.41 (0.01)$   &   $0.52 (0.01)$   &   $1.13 (0.00)$   &   $-0.29 (0.00)$   &   $0.36 (0.00)$   &   $0.81 (0.00)$   \\
\midrule
SGPR300   &   $0.01 (0.01)$   &   $0.42 (0.01)$   &   $\textbf{0.48 (0.01)}$   &   $\textbf{0.97 (0.01)}$   &   $\textbf{-0.45 (0.00)}$   &   $\textbf{0.16 (0.01)}$   &   $0.82 (0.00)$   \\
FGP300   &   $0.72 (0.04)$   &   $0.46 (0.01)$   &   $0.55 (0.01)$   &   $1.10 (0.01)$   &   $-0.37 (0.00)$   &   $0.40 (0.00)$   &   $0.82 (0.00)$   \\
MGP300   &   $\textbf{-0.01 (0.00)}$   &   $\textbf{0.40 (0.01)}$   &   $0.50 (0.01)$   &   $1.02 (0.01)$   &   $-0.30 (0.00)$   &   $0.21 (0.00)$   &   $\textbf{0.79 (0.00)}$   \\
\bottomrule
\end{tabular}
\end{sc}
\end{small}
\endgroup
\caption{Negative log-predictive density comparison (standard deviations reported in parentheses) on seven standard benchmark real-world datasets
The lowest negative log-predictive density is in bold.} 
\label{table:nlpd_time}
\end{table*}

\begin{figure*}[h]
\centering
\includegraphics[width=15.5cm,height=3cm]{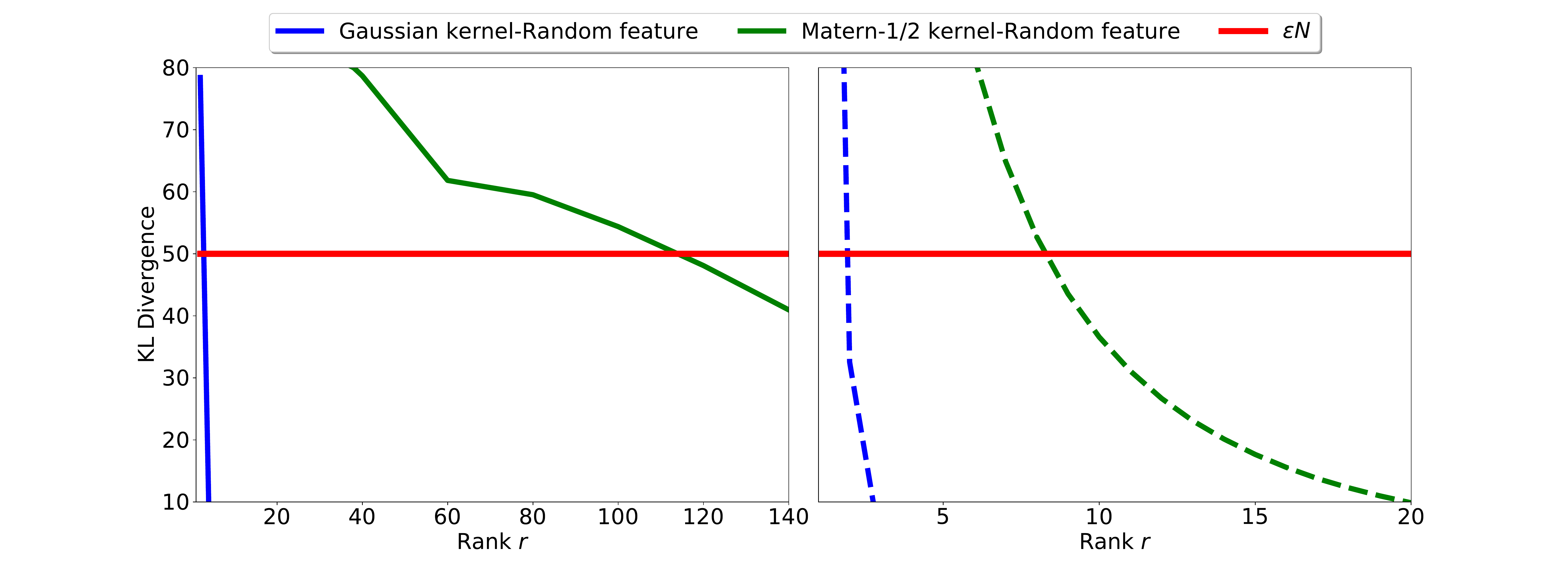} 
\caption{Decay of $\kld{\mathcal{N}(0, K + \sigma^2 I_N)}{\mathcal{N}(0,\Sigma + \sigma^2 I_N)}$ as the rank $r$ of $\Sigma$ increases, in the same settings as those in Fig \ref{fig:theorem_3_5_ranks} where $D=1$. Results from the random feature-based low-rank approximation (i.e.~setting of Thm~\ref{thm:KL divergence instantiation random fourier features}) and the Mercer expansion truncation-based low-rank approximation (i.e.~setting of Thm~\ref{thm:KL divergence instantiation mercer expansion}) are depicted on the left and right panel, respectively.} 
\label{fig:theorem_3_5_instance}
\end{figure*}

\subsection{Experimental rates of convergence}\label{sec:rates}

In this Section we use simulated data experiments to investigate the effectiveness of our theoretical bounds in capturing the dependence of the KL approximation on the dimension of the input features and the rank of $K$. 

Fig \ref{fig:theorem_3_5_ranks} depicts, for fixed $\varepsilon N$, the lowest rank $r$ needed for $\Sigma$ such that $\kld{\mathcal{N}(0, K + \sigma^2 I_N)}{\mathcal{N}(0,\Sigma + \sigma^2 I_N)} \leq \varepsilon N$ against the feature dimension $D$.  Our theoretical results are supported by comparing the blue against green and the solid against the dotted curves: the former indicates that the Gaussian kernels  require lower rank kernels to achieve the desired threshold  when compared with Mat\'ern kernels \footnote{Due to the limitation of obtaining closed-form expressions for the eigenvalues/eigenfunctions of Mat\'ern kernels, we have recourse to approximation of their values by the eigendecomposition of the data kernel matrix.} whereas the latter supports the theoretical results stating that the Mercer approximations  require lower rank kernels to achieve a desired threshold $\varepsilon N$ when compared with a random feature approximations.

Fig \ref{fig:theorem_3_5_instance} illustrates the practical behaviour, as a function of the rank, of the KL-divergence between an exact GP and an approximate GP obtained using random features or Mercer expansion truncation, in two example settings covered by Thms \ref{thm:KL divergence instantiation random fourier features} and 
\ref{thm:KL divergence instantiation mercer expansion}. Our theoretical bounds suggest that, for a fixed rank of $\Sigma$, approximating the Gaussian kernel using random features results in worst KL approximation compared to  approximating it by truncating its Mercer expansion. This is indeed reflected in our experiments on simulated data, when comparing the blue curves of the two panels of Fig~\ref{fig:theorem_3_5_instance}. Similarly, our theoretical bounds suggest that truncating the Mercer expansion of the Gaussian kernel provides better KL approximation compared to truncating the expansion of the Mat\'ern kernel, and this is indeed reflected when comparing the blue and green curves of the right panel of Fig~\ref{fig:theorem_3_5_instance}.

\subsection{Real data experiments}\label{sec:real data experiments}
We conduct a thorough experimental study, testing the quality of FGP and MGP over seven datasets from the UCI repository \cite{Dua2019UCI}. We also compare their performance to that of Sparse GP Regression (\emph{SGPR})  \citep{titsias2009variational} which   uses the Nystr\"{o}m approximation to conduct inference and whose convergence rates were studied by \citet{burt2019rates}. 

Both input data and their corresponding responses are standardized for all datasets. For Bike dataset, we use the standard dataset, however, we one-hot encoded some of the categorical features which led to an increased dimensionality. We train all methods for $300$ epochs using Adam  \citep{kingma2014adam}. All GPs use Gaussian kernels with separate length-scale per dimension. All results have been averaged over five random splits (90\% train, 10\% test). The last number $r$ in an acronym indicates that a method was trained using rank $r$. We use $r=6,10,50,100,200,300$ for all three methods.
For MGP, the projection dimension $d$ is determined by cross-validation on training data, with its value ranging in $3 \le d \le 7$ across all seven datasets. Table \ref{table:nlpd_time} presents comparisons of all methods in terms of negative log-predictive density (NLPD) where a similar table with root mean squared error (RMSE) scores is given in appendix. Table  \ref{table:nlpd_time} indicates that MGP has similar behavior with SGPR.  This is consistent with our theoretical predictions as our bounds for MGP and those of \citet{burt2019rates} for SGPR are quantitatively similar; see discussion in Sec~\ref{sec:related}.
The slight inferior performance of FGP compared to MGP is also consistent with the predictions of the theory as the bound of Thm~\ref{thm:KL divergence instantiation random fourier features} scales worse than that of Thm~\ref{thm:KL divergence instantiation mercer expansion}.


\section*{Acknowledgements}
C.D. and P.D. acknowledge partial financial support by the Alan Turing Institute under the EPSRC
grant EP/N510129/1. C.D. was supported by NSF Awards IIS-1741137, CCF-1617730 and CCF1901292, by a Simons Investigator Award, by the DOE PhILMs project (No. DE-AC05-76RL01830), and by the DARPA award HR00111990021. The authors would like to thank Andrew Ilyas for helping them with setting up the Linux machine used for the experiments.

\bibliography{main}

\clearpage

\appendix

\section{Omitted proofs}

\begin{prevproof}{Proposition~\ref{thm:KL approximation theorem}}
We first show~\eqref{eq:costas bound 1}. Recall that the KL divergence between two Gaussians with non-singular covariances has a closed form expression:
\begin{align}
    & \kld{\mathcal{N}(0,\Sigma_1)}{\mathcal{N}(0,\Sigma_2)} = \nonumber \\
    & {1 \over 2}\left({\rm Tr}(\Sigma_2^{-1}\Sigma_1) - N + \ln{|\Sigma_2| \over |\Sigma_1|}\right). \label{eq:KL bound}
\end{align}
Because $\Sigma_2$ is positive definite, $\Sigma_2^{-1}$ is too and it has a square root. Thus, by using properties of the trace we can write:
\begin{align*}
    &{\rm Tr}(\Sigma_2^{-1}\Sigma_1)={\rm Tr}(\Sigma_2^{-1/2}\Sigma_1\Sigma_2^{-1/2})\\
    &={\rm Tr}(\Sigma_2^{-1/2}(\Sigma_1-(1-\gamma)\Sigma_2 +(1-\gamma)\Sigma_2)\Sigma_2^{-1/2})\\
    &={\rm Tr}(\Sigma_2^{-1/2}(\Sigma_1-(1-\gamma)\Sigma_2 )\Sigma_2^{-1/2}) \\
    & ~~~~~+{\rm Tr}(\Sigma_2^{-1/2}((1-\gamma)\Sigma_2)\Sigma_2^{-1/2})\\
    &={\rm Tr}(\Sigma_2^{-1/2}(\Sigma_1-(1-\gamma)\Sigma_2)\Sigma_2^{-1/2})+(1-\gamma){\rm Tr}(I_N)\\
    &={\rm Tr}(\Sigma_2^{-1/2}(\Sigma_1-(1-\gamma)\Sigma_2 )\Sigma_2^{-1/2})+(1-\gamma)N
\end{align*}
Plugging this into~\eqref{eq:KL bound} yields:
\begin{align}
    &\kld{\mathcal{N}(0,\Sigma_1)}{\mathcal{N}(0,\Sigma_2)} = - {1 \over 2}\left(\gamma N - \ln{|\Sigma_2| \over |\Sigma_1|}\right) \nonumber \\
    & + {1 \over 2}\left({\rm Tr}(\Sigma_2^{-1/2}(\Sigma_1-(1-\gamma)\Sigma_2 )\Sigma_2^{-1/2})\right). \label{eq:KL bound dervitation stop 1}
\end{align}
Next we argue the following:
\begin{lemma} \label{lem:helper 1}
If $A, B$ are positive definite, and $B - A$ is positive semidefinite, then $\ln\left({|A| \over |B|} \right)\le 0$.
\end{lemma}

\begin{prevproof}{Lemma~\ref{lem:helper 1}}
Let $\la_1 \ge \la_2 \ge \ldots \ge\la_N >0$ be the eigenvalues of $A$, and $\la_1' \ge \la_2' \ge \ldots \ge \la_N' >0$ be the eigenvalues of $B$,  in non-increasing order. Because $B \succeq A$, by the min-max theorem \cite{fan1953minimax} we have $\la_i \le \la'_i$,  $\forall i$. Thus,
\begin{align*}
    {|A|\over |B|}= {\prod_{i=1}^N {\la_i \over \la'_i} \le 1} \Rightarrow \ln\left({|A|\over |B|}\right) \le 0.
\end{align*}
\end{prevproof}
Because $(1+\gamma)\Sigma_1 \succeq \Sigma_2$, it follows from Lemma~\ref{lem:helper 1} that
\begin{align*}
    0 &\ge \ln\left({|\Sigma_2| \over |(1+\gamma)\Sigma_1|} \right)
    = \ln\left({|\Sigma_2| \over (1+\gamma)^N |\Sigma_1|}\right) \\
    & = \ln\left({|\Sigma_2| \over |\Sigma_1|}\right) - N \ln(1+\gamma)
    \ge \ln\left({|\Sigma_2| \over |\Sigma_1|}\right) - N \gamma.
\end{align*}
Combining the last inequality with~\eqref{eq:KL bound dervitation stop 1} yields Bound~\eqref{eq:costas bound 1}.

\bigskip To prove~\eqref{eq:costas bound 2}, we note that if additionally $(1+\gamma)\Sigma_2 \succeq \Sigma_1$ then:
\begin{align}
    {1 \over 2}{\rm Tr}(\Sigma_2^{-1/2}((1+\gamma)\Sigma_2-\Sigma_1)\Sigma_2^{-1/2})\ge 0.\label{eq:KL bound derivation stop 1.5}
\end{align}
This follows by noticing that matrix $\Sigma_2^{-1/2}((1+\gamma)\Sigma_2-\Sigma_1)\Sigma_2^{-1/2} \succeq 0$. Indeed, for all $\bx\in \mathbb{R}^N$ and using that $(\Sigma_2^{-1/2})^{\rm T}=\Sigma_2^{-1/2}$:
\begin{align*}
    & \bx^{\rm T} \Sigma_2^{-1/2}((1+\gamma)\Sigma_2-\Sigma_1)\Sigma_2^{-1/2} \bx \\
    & =(\Sigma_2^{-1/2} \bx)^{\rm T}((1+\gamma)\Sigma_2-\Sigma_1)(\Sigma_2^{-1/2} \bx) \ge 0,
\end{align*}
where the last inequality follows from the positive semidefiniteness of $(1+\gamma)\Sigma_2-\Sigma_1$.

Now combining~\eqref{eq:KL bound derivation stop 1.5} with~\eqref{eq:costas bound 1} and using properties of the trace we get:
\begin{align*}
    &\kld{\mathcal{N}(0,\Sigma_1)}{\mathcal{N}(0,\Sigma_2)}\\ 
    &~~~~~~\le {1 \over 2} \Biggl( {\rm Tr}(\Sigma_2^{-1/2}(\Sigma_1-(1-\gamma)\Sigma_2)\Sigma_2^{-1/2}) \\
    &~~~~~~~~~~~+ {\rm Tr}(\Sigma_2^{-1/2}((1+\gamma)\Sigma_2-\Sigma_1)\Sigma_2^{-1/2})\Biggr)\\
    &~~~~~~\le {1 \over 2} \left( {\rm Tr}(\Sigma_2^{-1/2}(2\gamma\Sigma_2)\Sigma_2^{-1/2}) \right)\\
    &~~~~~~\le \gamma {\rm Tr}(I_N) = \gamma N.
\end{align*}

\bigskip Let us now move to the proof of~\eqref{eq:costas bound}. We plug $\Sigma_1=\sigma^2 I_N + K_1$ and $\Sigma_2=\sigma^2I_N+K_2$ into~\eqref{eq:costas bound 1} to get:
\begin{align}
    & \kld{\mathcal{N}(0,\Sigma_1)}{\mathcal{N}(0,\Sigma_2)} \nonumber\\
    &\le {1 \over 2}{\rm Tr}(\Sigma_2^{-1/2}(K_1-(1-\gamma)K_2+\gamma \sigma^2 I_N)\Sigma_2^{-1/2}) \nonumber\\
    &\le {1 \over 2}{\rm Tr}(\Sigma_2^{-1}(K_1-(1-\gamma)K_2+\gamma \sigma^2 I_N))\label{eq:KL bounds stop 2}
\end{align}
where we used properties of the trace. Because $K_2$ is positive semidefinite, it has eigenvalues $\la_1\ge \la_2 \ge \ldots \ge \la_N\ge 0$, which implies that $\Sigma_2=\sigma^2 I + K_2$ has eigenvalues $\sigma^2+\la_1\ge \sigma^2+\la_2 \ge \ldots \ge \sigma^2+ \la_N> 0$, which in turn implies that $\Sigma_2^{-1}$ has eigenvalues $(\sigma^2+\la_N)^{-1}\ge (\sigma^2+\la_{N-1})^{-1} \ge \ldots \ge (\sigma^2+ \la_1)^{-1}> 0$. Now using~\eqref{eq:KL bounds stop 2} and properties of the trace we have that:
\begin{align*}
& \kld{\mathcal{N}(0,\Sigma_1)}{\mathcal{N}(0,\Sigma_2)} \\
& \le {1 \over 2}{\rm Tr}(\Sigma_2^{-1}(K_1-(1-\gamma)K_2+\gamma \sigma^2 I_N))\\&\le {1 \over 2}\lambda_{\max}(\Sigma_2^{-1}){\rm Tr}(K_1-(1-\gamma)K_2+\gamma \sigma^2 I_N)\\
    &= {1 \over 2}\cdot {1 \over \sigma^2+\la_N}\cdot {\rm Tr}(K_1-(1-\gamma)K_2+\gamma \sigma^2 I_N)\\
    &\le {1 \over 2 \sigma^2}{\rm Tr}(K_1-(1-\gamma)K_2+\gamma \sigma^2 I_N),
\end{align*}
where in the above derivation $\lambda_{\max}(\Sigma_2^{-1})$ is the maximum eigenvalue of matrix $\Sigma_2^{-1}$.
\end{prevproof}

\begin{prevproof}{Theorem~\ref{thm:KL divergence instantiation random fourier features}} We will make use of the following theorem.

\begin{theorem}[Theorem 12 of~\cite{avron2017random}] \label{thm:spectral approximation of Gaussian kernel}
Consider the $D$-dimensional Gaussian kernel $k(\bx,\bx')=\exp(-2\pi^2||\bx-\bx'||_2^2)$, and the kernel matrix $K=K(k,X)=(k(\bx_i,\bx_j))_{ij}$, where $X=(\bx_1,\ldots,\bx_N)$ is a collection of points in $\mathbb{R}^D$ such that, for some $R>0$, $||\bx_i-\bx_j||_{ \infty }\le R, \forall i,j$. Suppose $D \le 5 \log(N/\sigma^2)+1$ and $\varepsilon \in (0,1)$. There exists (a samplable in $O(D)$-time) distribution $p(\boldeta)$ and a parameterized family $e_{\boldeta}(\cdot)$ of {\em modified} Fourier Features such that, if $r \ge \Omega\left({R^D \over \varepsilon^2} \left(\log{N \over \sigma^2}\right)^{2D} \log\left({s_{\sigma^2}(K)\over\delta}\right)\right)$, where $s_{\sigma^2}(K)={\rm Tr}((\sigma^2I+K)^{-1}K)$ and $\delta\in(0,1)$, then the feature map $\phi(\bx)=(e_{\boldeta_1}(\bx),\ldots,e_{\boldeta_r}(\bx))$ where $\boldeta_1,\ldots,\boldeta_r \sim p(\boldeta)$ satisfies the following with probability at least $1-\delta$:
\begin{align}
 (1-\varepsilon)(\sigma^2 I_N + K) \preceq  (\sigma^2 I_N + \Sigma) \preceq
 (1+\varepsilon)(\sigma^2 I_N + K),
\end{align}
where $\Sigma=(\phi(\bx_i)^{\rm \top} \phi(\bx_j))_{ij}$, and $\preceq$ denotes semi-definite domination.
\end{theorem}

Now set $\Sigma_1 = \sigma^2 I_N + K$ and $\Sigma_2= \sigma^2 I_N + \Sigma$. Notice that $s_{\sigma^2}(K)={\rm Tr}((\sigma^2I+K)^{-1}K) \le {\rm Tr}(I_N) \le N$. Thus, given our choice of $r$, Theorem~\ref{thm:spectral approximation of Gaussian kernel} implies that, with probability at least $1-\delta$, $\Sigma_1$ and $\Sigma_2$ satisfy:
\begin{align*}
(1-\varepsilon)\Sigma_1 \preceq \Sigma_2 \preceq (1+\varepsilon)\Sigma_1.
\end{align*}
Given that for $\varepsilon \in (0,{1\over 2}]$, we get that $1-\varepsilon \ge {1 \over 1+2\varepsilon}$, the above implies that:
\begin{align*}
(1+2\varepsilon)^{-1}\Sigma_1 \preceq \Sigma_2 \preceq (1+2\varepsilon)\Sigma_1.
\end{align*}
Now we use~\eqref{eq:costas bound 2} of Proposition~\ref{thm:KL approximation theorem}, to get that the KL divergence from distribution~\eqref{eq:approximate model} to distribution~\eqref{eq:idealized model}  is bounded by $2\varepsilon N$.
\end{prevproof}

\begin{prevproof}{Theorem~\ref{thm:KL divergence instantiation mercer expansion}}
We will make use of the following theorem:
\begin{theorem}[Proof of Theorem 4 in~\cite{braun2006accurate}] \label{thm:trace bound Mercer truncation}
Let $k(\cdot,\cdot)$ be a Mercer kernel on probability space $({\cal X},\mu)$ with $k(\bx,\bx)\le B$, for all $\bx \in {\cal X}$. Let $X=(\bx_1,\ldots,\bx_N)$ comprise samples from $\mu$, let $K=K(k,X)$ (which satisfies~\eqref{eq:expansion of kernel matrix}), and let $\Sigma=\sum_{\kk=1}^{r} \lambda_{\kk} \xii_{\kk} \xii_{\kk}^{\top}$, for some $r \in \mathbb{N}$ (which has rank $r$). With probability at least $1-\delta$ over the samples $X$:
\begin{align}
    {\rm Tr}(K-\Sigma) \le N \cdot \left( \Lambda_{>r} + \sqrt{{B\Lambda_{>r} \over N\delta} }\right), \label{eq:good trace bound event}
\end{align}
where $\Lambda_{>r}=\sum_{\kk>r}\lambda_\kk$.
\end{theorem}

To prove the first part of our theorem, notice that, because $\Sigma$ is a truncation of $K$, $K-\Sigma$ is positive semidefinite. To prove~\eqref{eq:KL bound from mercer truncation}, we set $K_1=K$, $K_2=\Sigma$, and use~\eqref{eq:costas bound} from Proposition~\ref{thm:KL approximation theorem} with $\gamma=0$ to get that the KL divergence from distribution~\eqref{eq:approximate model} to distribution~\eqref{eq:idealized model} is bounded by:
\begin{align*}
  {1 \over 2\sigma^2} {\rm Tr}(K-\Sigma) \overset{\eqref{eq:good trace bound event}}{\le} {N \over 2\sigma^2} \cdot \left( \Lambda_{>r} + \sqrt{{B\Lambda_{>r} \over N \delta} }\right).  
\end{align*}
    
Next, we prove our bound for the Gaussian kernel using properties of its eigenspectrum. This is well-understood; see e.g.~\cite{williams2006gaussian}. For completeness we also describe it in Section~\ref{sec:gaussian kernel mercer illustration}. As per Equations~\eqref{kernel}, \eqref{eigen}, \eqref{eigenvalues} in that section, the eigenfunctions and eigenvalues of the Gaussian kernel can be indexed by vectors $\bn \in \mathbb{N}^d$. Let us pick an arbitrary, absolute constant $0<c<1$ and split our analysis into two cases: $R \ge c$ and $R \le c$. 
\begin{itemize}

\item Case $R \ge c$: Plugging into~\eqref{eigenvalues} and simple manipulations, we obtain that the eigenvalues satisfy $\lambda_{\bn} \le \left({R+1 \over R^2}\right)^D {\left(1-{1 \over R+1} \right)^{\mathbbm{1}
^{\rm T} \bn} }$, where  $\mathbbm{1}$ denotes the vector of all ones. Moreover, the eigenvalues are ordered in terms of the ``level sets'' of $\mathbbm{1}^{\rm T} \bn$; in particular, the larger $\mathbbm{1}
^{\rm T} \bn$ is, the smaller the eigenvalue is, while every $\bn$ with the same value of $\mathbbm{1}^{\rm T} \bn$ has the same eigenvalue. For $m = \Omega( RD \log (RD) + R\log{1 \over \varepsilon \sigma \delta})$, let us take $r=|\{\bn \in \mathbb{N}^D~|~\mathbbm{1}^{\rm T} \bn < m \}|$. We have that
\begin{align}
&\Lambda_{>r}=\sum_{\bn: \mathbbm{1}^{\rm T} \bn \ge m} \lambda_{\bn} \notag\\ 
&\le \sum_{\bn: \mathbbm{1}^{\rm T} \bn \ge m} \left({R+1 \over R^2}\right)^D {\left(1-{1 \over R+1} \right)^{\mathbbm{1}
^{\rm T} \bn} }\notag\\
&\le \sum_{\ell = m}^{\infty} \ell^D \left({R+1 \over R^2}\right)^D {\left(1-{1 \over R+1} \right)^{\ell} }\notag\\
&= \left({R+1 \over R^2}\right)^D \sum_{\ell = m}^{\infty} \left(\ell^D \left({R \over R+1}\right)^{\ell/2}\right) \left({R \over R+1}\right)^{\ell/2}\notag\\
&\le \left({R+1 \over R^2}\right)^D \sum_{\ell = m}^{\infty}\left({R \over R+1}\right)^{\ell/2}\notag\\
&\le \left({R+1 \over R^2}\right)^D \left({R \over R+1}\right)^{m/2} \cdot {1 \over 1- \sqrt{{R \over R+1}}} \notag\\
&= \left({R+1 \over R^{2-1/D}}\right)^D \left({R \over R+1}\right)^{m/2} \cdot {1 \over R \left(1- \sqrt{{R \over R+1}}\right)}, \label{eq:final kourasi}
\end{align}
where the second to last inequality follows from the fact that $\ell^D \left({R \over R+1}\right)^{\ell/2} \le 1$ for $\ell \ge m=\Omega(RD \log (RD \vee e))$. To conclude the proof notice that the Gaussian kernel $k(\bx,\bx')=\exp(-2\pi^2||\bx-\bx'||_2^2)$ satisfies $k(\bx,\bx)=1$, hence we can use~\eqref{eq:KL bound from mercer truncation} with $B=1$ to bound the KL divergence from distribution~\eqref{eq:approximate model} to distribution~\eqref{eq:idealized model} by
\begin{align}
{N \over 2\sigma^2} \cdot \left( \Lambda_{>r} + \sqrt{{\Lambda_{>r} \over N \delta} }\right) \le \varepsilon N, \label{eq: final kourasi 3}
\end{align}
where the last inequality uses~\eqref{eq:final kourasi} and that $m = \Omega(RD \log (RD \vee e) + R\log{1 \over \varepsilon \sigma \delta})$. Given that $r=|\{\bn \in \mathbb{N}^D~|~\mathbbm{1}^{\rm T} \bn < m \}|$, we get that to attain~\eqref{eq: final kourasi 3} it suffices to choose the rank to be $r=m^D=\left(\Omega(RD \log (RD \vee e) + R\log{1 \over \varepsilon \sigma \delta})\right)^D$.

\item Case $R \le c$: Plugging into~\eqref{eigenvalues} and simple manipulations, we obtain that the eigenvalues satisfy $\lambda_{\bn} \le \left({\sqrt{R^2+1} \over R^2}\right)^D {\left(1-{1 \over R^2+1} \right)^{\mathbbm{1}
^{\rm T} \bn} }$, where  $\mathbbm{1}$ denotes the vector of all ones. Moreover, the eigenvalues are ordered in terms of the ``level sets'' of $\mathbbm{1}^{\rm T} \bn$; in particular, the larger $\mathbbm{1}
^{\rm T} \bn$ is, the smaller the eigenvalue is, while every $\bn$ with the same value of $\mathbbm{1}^{\rm T} \bn$ has the same eigenvalue. For $m$ equals to
\begin{align*}
    \Omega \Biggl( {D \over \log(1+{1 \over R^2})} & \left(\log{D \over \log(1+{1 \over R^2})} \vee \log{\sqrt{R^2+1} \over R^2} \right)  \\
    & + {1 \over \log(1+{1 \over R^2})}\log{1 \over \varepsilon \sigma \delta} \Biggr),
\end{align*}
 let us take $r=|\{\bn \in \mathbb{N}^D~|~\mathbbm{1}^{\rm T} \bn < m \}|$. We have that
\begin{align}
&\Lambda_{>r}=\sum_{\bn: \mathbbm{1}^{\rm T} \bn \ge m} \lambda_{\bn} \notag\\ 
&\le \sum_{\bn: \mathbbm{1}^{\rm T} \bn \ge m} \left({\sqrt{R^2+1} \over R^2}\right)^D {\left(1-{1 \over R^2+1} \right)^{\mathbbm{1}
^{\rm T} \bn} }\notag\\
&\le \sum_{\ell = m}^{\infty} \ell^D \left({\sqrt{R^2+1} \over R^2}\right)^D {\left(1-{1 \over R^2+1} \right)^{\ell} }\notag\\
&= \left({\sqrt{R^2+1} \over R^2}\right)^D \sum_{\ell = m}^{\infty} \left(\ell^D \left({R^2 \over R^2+1}\right)^{\ell/2}\right) \left({R^2 \over R^2+1}\right)^{\ell/2}\notag\\
&\le \left({\sqrt{R^2+1} \over R^2}\right)^D \sum_{\ell = m}^{\infty}\left({R^2 \over R^2+1}\right)^{\ell/2}\notag\\
&\le \left({\sqrt{R^2+1} \over R^2}\right)^D \left({R^2 \over R^2+1}\right)^{m/2} \cdot {1 \over 1- \sqrt{{R^2 \over R^2+1}}}  \label{eq:final kourasi for R small}
\end{align}
where the second to last inequality follows from the fact that $\ell^D \left({R^2 \over R^2+1}\right)^{\ell/2} \le 1$ for $\ell \ge m=\Omega\left( {D \over \log(1+{1 \over R^2})} \log \left({D \over \log(1+{1 \over R^2})} \vee e\right) \right)$. To conclude the proof notice that the Gaussian kernel $k(\bx,\bx')=\exp(-2\pi^2||\bx-\bx'||_2^2)$ satisfies $k(\bx,\bx)=1$, hence we can use~\eqref{eq:KL bound from mercer truncation} with $B=1$ to bound the KL divergence from distribution~\eqref{eq:approximate model} to distribution~\eqref{eq:idealized model} by
\begin{align}
{N \over 2\sigma^2} \cdot \left( \Lambda_{>r} + \sqrt{{\Lambda_{>r} \over N \delta} }\right) \le \varepsilon N, \label{eq: final kourasi 2}
\end{align}
where the last inequality uses~\eqref{eq:final kourasi for R small} and 
\begin{align*}
m = \Omega \Biggl( &{D \over \log(1+{1 \over R^2})} \log{\sqrt{R^2+1} \over R^2} \\
&+ {1 \over \log(1+{1 \over R^2})}\log{1 \over \varepsilon \sigma \delta}\Biggr).
\end{align*} 
Given that $r=|\{\bn \in \mathbb{N}^D~|~\mathbbm{1}^{\rm T} \bn < m \}|$, we get that to attain~\eqref{eq: final kourasi 2} it suffices to choose the rank to be \begin{align*}
&r=m^D= \\
&\Biggl(\Omega \Biggl( {D \over \log(1+{1 \over R^2})} \left(\log{D \over \log(1+{1 \over R^2})} \vee \log{\sqrt{R^2+1} \over R^2} \right) \\
&\quad \quad \quad \quad \quad \quad \quad \quad \quad \quad + {1 \over \log(1+{1 \over R^2})}\log{1 \over \varepsilon \sigma \delta}\Biggr) \Biggr)^D.
\end{align*} 
Given the above, taking $r$ as follows suffices (which might be a more convenient form of a sufficient bound):
\begin{align*}
r=\Biggl(\Omega\Biggl( {D \over \log{1 \over R}} &\left(\log{D \over \log{1 \over R}} \vee \log{2 \over R^2} \right) \\
&~~~~+{1 \over \log{1 \over R}}\log{1 \over \varepsilon \sigma \delta}\Biggr)\Biggr)^D.
\end{align*}
\end{itemize}

Finally, we prove our bound for the Mat\'ern kernel with parameter $\nu$ and length scale $\alpha$. It follows from~\cite{seeger2008information} that for some constants $C$ and $s_0$ that depend on $D, \nu, \alpha$ and the bounds on $\mu$ and its support, the eigenvalues $\lambda_1 \ge \lambda_2 \ge \ldots$ of the kernel with respect to measure $\mu$ satisfy that
\begin{align*}\lambda_m \le C \left({1 \over m}\right)^{2\nu + D \over D}, \forall m \ge s_0.\end{align*}
It follows that for any $r\ge s_0$, we have
\begin{align}
    \Lambda_{>r} & = C\sum_{m \ge r+1}\left({1 \over m}\right)^{2\nu+D \over D} \nonumber\\
    & \le C\int_{r}^{+\infty}{1 \over x^{2\nu+D \over D}}dx = C{D \over 2\nu} {1 \over r^{2\nu \over D}}. \label{eq:matern kourasoula}
\end{align}
To conclude the proof notice that the Mat\'ern kernel, as stated in the statement of the theorem, satisfies $k(\bx,\bx)=1$. Hence, we can use~\eqref{eq:KL bound from mercer truncation} with $B=1$ to bound the KL divergence from distribution~\eqref{eq:approximate model} to distribution~\eqref{eq:idealized model} by
\begin{align}
{N \over 2\sigma^2} \cdot \left( \Lambda_{>r} + \sqrt{{\Lambda_{>r} \over N \delta} }\right) \le \varepsilon N, \label{eq: final kourasoula 2}
\end{align}
where the last inequality uses~\eqref{eq:matern kourasoula} and choosing $r \ge s_0 \vee A \left({1 \over \varepsilon \sigma \delta} \right)^{\Omega(D/\nu)}$ for some constant
 $A$ that depends on $\nu, \alpha, D$ and the bounds on $\mu$ and its support, but does not depend on $\varepsilon,\sigma,\delta,N$.
\end{prevproof}

\begin{prevproof}{Proposition~\ref{prop:from KL to parameter bounds}}
The KL divergence between two Gaussians has an explicit form:
\begin{align}
    &\kld{{\cal N}(\bmu_1,\Sigma_1)}{{\cal N}(\bmu_2,\Sigma_2)}=\notag\\&~~~~~{1 \over 2}\Biggl({\rm tr}(\Sigma_2^{-1}\Sigma_1) - N + \ln\left({|\Sigma_2| \over |\Sigma_1|}\right)\notag\\ &~~~~~~~~~~~~~+ (\bmu_2-\bmu_1)^{\rm T}\Sigma_2^{-1}(\bmu_2-\bmu_1) \Biggl). \label{eq:KL between two Gaussianssss}
\end{align}
Since $\Sigma_1, \Sigma_2$ are positive definite the following matrix is well-defined and positive definite as well: 
$$\Lambda = \Sigma_2^{-1/2}\Sigma_1 \Sigma_2^{-1/2}.$$
Moreover, observe that
\begin{align*}{\rm tr}(\Sigma_2^{-1}\Sigma_1) - N + \ln\left({|\Sigma_2| \over |\Sigma_1|}\right) &= {\rm tr}(\Lambda) - N - \ln\left({|\Lambda|}\right)\\
&= \sum_{i=1}^N \left(\lambda_i - 1 - \ln(\lambda_i)\right),
\end{align*}
where $0<\lambda_1,\ldots,\lambda_N$ are the eigenvalues of $\Lambda$. Plugging the above into~\eqref{eq:KL between two Gaussianssss} we get
\begin{align}
    &\kld{{\cal N}(\bmu_1,\Sigma_1)}{{\cal N}(\bmu_2,\Sigma_2)}=\notag\\&~~~~~{1 \over 2}\Biggl(\sum_{i=1}^N \left(\lambda_i - 1 - \ln(\lambda_i)\right)\notag\\ &~~~~~~~~~~~~~+ (\bmu_2-\bmu_1)^{\rm T}\Sigma_2^{-1}(\bmu_2-\bmu_1) \Biggl). \label{eq:KL between two Gaussianssss 2}
\end{align}
Next we observe that the function $x-1-\ln(x)\ge 0$, for all $x>0$. We also observe that, because $\Sigma_2^{-1}$ is positive definite, $(\bmu_2-\bmu_1)^{\rm T}\Sigma_2^{-1}(\bmu_2-\bmu_1) \ge 0$. These observations together with the hypothesis that $\kld{{\cal N}(\bmu_1,\Sigma_1)}{{\cal N}(\bmu_2,\Sigma_2)} \le \gamma$ from the proposition statement imply that:
\begin{align}
    {1 \over 2}(\bmu_2-\bmu_1)^{\rm T}\Sigma_2^{-1}(\bmu_2-\bmu_1) \le \gamma; \label{eq:first ask of the proposition}\\
    \lambda_i-1-\ln(\lambda_i)\le 2\gamma, \forall i. \label{eq: towards second ask of the proposition}
\end{align}
\eqref{eq:first ask of the proposition} is identical to~\eqref{eq:costas mean bounds} in the proposition statement. To show~\eqref{eq:costas cov bounds} we first observe that the function $x-1-\ln(x)$ is convex in its domain $x\in (0,+\infty)$ and attains its global minimum of $0$ at $x=1$. The equation $x-1-\ln(x)=2\gamma$ has thus exactly two roots $b(2\gamma)$ and $t(2\gamma)$ satisfying $0 < b(2\gamma) <1<t(2\gamma)$. Thus~\eqref{eq: towards second ask of the proposition} implies that $b(2\gamma) \le \lambda_i \le t(2\gamma),$ for all $i$. As $\lambda_1,\ldots,\lambda_N$ are the eigenvalues of $\Lambda = \Sigma_2^{-1/2}\Sigma_1 \Sigma_2^{-1/2}$ this implies that
$$b(2\gamma) \cdot \Sigma_2 \preceq \Sigma_1 \preceq t(2\gamma) \cdot \Sigma_2,$$
as we had wanted to show. Finally, it is easy to show using basic calculus that $b(2\gamma) \ge \max(1-2\sqrt{\gamma}, \exp(-1-2\gamma))$ and $t(2\gamma)\le 1+\max(\sqrt{8\gamma},8\gamma)$.
\end{prevproof}

\section{Mercer Expansion of the Multi-Dimensional Gaussian Kernel} \label{sec:gaussian kernel mercer illustration}

For illustration purposes we provide the Mercer expansion of the multi-dimensional Gaussian kernel, showing how its eigenvalue sequence decays exponentially fast. We consider a general $D$-dimensional Gaussian kernel as follows:
\begin{equation}
k_{\sigma_f^2,\Delta}(\bx_i,\bx_j) = \sigma^2_f \exp (- (\bx_i-\bx_j)^{\rm T} \Delta (\bx_i-\bx_j)), \label{eq:ard_kernel}
\end{equation}
where $\Delta = {\rm diag}(\epsilon_1^2,\ldots,\epsilon_D^2)$ contains the length scales along the $D$  dimensions of the covariates, and $\sigma^2_f$ is the variance. The parameters of the kernel are $\theta=(\sigma_f^2,\Delta)$.

We view $k_{\sigma_f^2,\Delta}(\bx_i,\bx_j)$ as a kernel over $\mathbb{R}^D$ equipped with an axis aligned Gaussian measure $\rho(\bx)=\rho(x^1,\ldots,x^D)$, whose density in dimension $j$ is given by
\begin{equation}
\rho_j(x^j) = \alpha_j \pi^{-1/2} \exp(-\alpha_j^2 (x^j)^2),~\forall j=1,\ldots,D. \label{input}
\end{equation}
Mercer's expansion theorem \cite{mercer1909functions} allows us to write 
\begin{equation}
k_{\sigma_f^2,\Delta}(\bx_i,\bx_j) = \sum_{\bn \in \mathbb{N}^D} \lambda_{\bn} \phii_{\bn}(\bx_i) \phii_{\bn}(\bx_j), \label{kernel}
\end{equation}
where $(\phii_{\bn})_{\bn \in \mathbb{N}^d}$ is an  orthonormal basis of $L_2(\mathbb{R}^D,\rho)$, wherein inner products are computed using $\rho(\bx)$. It is well-known
~\cite{williams2006gaussian,fasshauer2012stable,fasshauer2012green} that such an orthonormal basis $(\phii_{\bn})_{\bn \in \mathbb{N}^D}$ can be constructed as a tensor product of the orthonormal bases of $L_2(\mathbb{R}^D,\rho_j)$ for all $j$, as follows. Setting
$\beta_j  =  \left(   1 + (2\epsilon_j/\alpha_j)^2   \right)^{1/4}, \gamma_{n_j} = \beta_j^{1/2} 2^{(1-n_j)/2}\Gamma(n_j)^{-1/2}$ and
 $\delta_j^2  =  \alpha_j^2(\beta_j^2 -1)/2$
the orthonormal eigenvectors are defined as 
\begin{align}
\phii_{\bn}(\bx) & = \prod_{j=1}^D \phii_{n_j}(x^j) \nonumber \\
  & = \prod_{j=1}^D \gamma_{n_j} \exp(-\delta_j^2 (x^j)^2) H_{n_j-1} (\alpha_j \beta_j x^j) , \label{eigen}
\end{align}
where $H_n$ are the Hermite polynomials of degree $n$ and the corresponding eigenvalues are \\
\begin{align}
\lambda_{\bn} &= \sigma_f^2 \prod_{j=1}^{D} \lambda_{n_j} \nonumber \\
&=  \sigma_f^2 \prod_{j=1}^{D} 
\left(   \frac{\alpha_j^2}{\alpha_j^2+\delta_j^2+\epsilon_j^2}  \right)^{1/2} \left(   \frac{\epsilon_j^2}{\alpha_j^2+\delta_j^2+\epsilon_j^2}  \right)^{n_j-1} 
. \label{eigenvalues}
\end{align}
Note that $\lambda_{n_j} \rightarrow 0$ as $n_j \rightarrow \infty$. Indeed, as long as  $\alpha_j^2/\epsilon_j^2$ is bounded away from $0$, this decay is exponentially fast.

%

\section{Additional experimental results}\label{sec:extra_expers}

\subsection{Extra results on real data }\label{sub_sec:rmse_real}

Table~\ref{table:rmse} demonstrates the RMSE values of all methods discussed in Section~\ref{sec:real data experiments}. RMSE values follow similar trends as the corresponding NLPD values reported in Table~\ref{table:nlpd_time}.

\begin{table*}[h!]
\begin{center}
\begingroup
\setlength{\tabcolsep}{9.7pt} 
\begin{small}
\begin{sc}
\begin{tabular}{l c c c c c c c}
\toprule
&  \multicolumn{6}{c}{\textbf{RMSE}}  \\
\cmidrule(lr){2-8}
   &   Bike & Elevators   &   Super & Protein   &   Sarcos   &   KeggDir & 3DRoad      \\
$N_{\rm train}$ &   $ 15641$   &   $14939 $   &   $ 19136 $   &   $41157$   &   $44039$   &   $48071 $   &   $391386$   \\
$N_{\rm test}$ &  $1738$ & $1660 $   &   $2127 $   &   $ 4573 $   &   $ 4894 $   &   $5342 $   &   $43488$   \\
$D$ &  $57$ & $ 18 $   &   $81$   &   $ 9 $   &   $ 21 $   &   $19 $   &   $3 $    \\
\midrule
SGPR6   &   $1.01 (0.01)$   &   $0.40 (0.01)$   &   $0.47 (0.00)$   &   $0.82 (0.00)$   &   $0.33 (0.01)$   &   $0.46 (0.01)$   &   $0.90 (0.01)$   \\
FGP6   &   $0.61 (0.02)$   &   $0.77 (0.18)$   &   $0.52 (0.01)$   &   $0.88 (0.01)$   &   $0.36 (0.00)$   &   $0.86 (0.01)$   &   $0.87 (0.00)$   \\
MGP6   &   $0.33 (0.01)$   &   $0.42 (0.01)$   &   $0.45 (0.01)$   &   $0.82 (0.00)$   &   $0.24 (0.00)$   &   $0.54 (0.00)$   &   $0.90 (0.01)$   \\
\midrule
SGPR10   &   $0.65 (0.28)$   &   $0.40 (0.01)$   &   $0.45 (0.01)$   &   $0.79 (0.00)$   &   $0.28 (0.02)$   &   $0.43 (0.01)$   &   $0.70 (0.00)$   \\
FGP10   &   $0.60 (0.01)$   &   $0.81 (0.24)$   &   $0.52 (0.00)$   &   $0.82 (0.01)$   &   $0.34 (0.00)$   &   $0.70 (0.01)$   &   $0.81 (0.00)$   \\
MGP10   &   $0.27 (0.01)$   &   $0.40 (0.01)$   &   $0.40 (0.01)$   &   $0.77 (0.00)$   &   $0.21 (0.00)$   &   $0.47 (0.00)$   &   $0.75 (0.01)$   \\
\midrule
SGPR50   &   $0.25 (0.01)$   &   $0.38 (0.01)$   &   $0.42 (0.01)$   &   $0.73 (0.01)$   &   $0.19 (0.00)$   &   $0.34 (0.01)$   &   $0.61 (0.00)$   \\
FGP50   &   $0.60 (0.01)$   &   $0.42 (0.01)$   &   $0.48 (0.01)$   &   $0.79 (0.00)$   &   $0.22 (0.00)$   &   $0.49 (0.01)$   &   $0.70 (0.00)$   \\
MGP50   &   $0.26 (0.01)$   &   $0.39 (0.01)$   &   $0.39 (0.01)$   &   $0.75 (0.00)$   &   $0.18 (0.00)$   &   $0.38 (0.00)$   &   $0.65 (0.00)$   \\
\midrule
SGPR100   &   $0.25 (0.01)$   &   $0.38 (0.01)$   &   $0.40 (0.01)$   &   $0.70 (0.01)$   &   $0.18 (0.00)$   &   $0.32 (0.00)$   &   $0.57 (0.01)$   \\
FGP100   &   $0.55 (0.04)$   &   $0.41 (0.01)$   &   $0.46 (0.00)$   &   $0.77 (0.00)$   &   $0.20 (0.00)$   &   $0.44 (0.01)$   &   $0.66 (0.00)$   \\
MGP100   &   $0.25 (0.01)$   &   $0.38 (0.01)$   &   $0.38 (0.01)$   &   $0.74 (0.00)$   &   $0.18 (0.00)$   &   $0.35 (0.00)$   &   $0.59 (0.00)$   \\
\midrule
SGPR200   &   $0.24 (0.01)$   &   $0.38 (0.01)$   &   $0.39 (0.01)$   &   $0.66 (0.01)$   &   $0.16 (0.00)$   &   $0.30 (0.01)$   &   $0.56 (0.01)$   \\
FGP200   &   $0.53 (0.03)$   &   $0.39 (0.01)$   &   $0.43 (0.01)$   &   $0.75 (0.01)$   &   $0.18 (0.00)$   &   $0.39 (0.01)$   &   $0.59 (0.00)$   \\
MGP200   &   $0.24 (0.01)$   &   $0.37 (0.01)$   &   $0.38 (0.01)$   &   $0.73 (0.00)$   &   $0.17 (0.00)$   &   $0.34 (0.00)$   &   $0.55 (0.00)$   \\
\midrule
SGPR300   &   $\textbf{0.24 (0.01)}$   &   $\textbf{0.37 (0.01)}$   &   $\textbf{0.38 (0.01)}$   &   $\textbf{0.64 (0.01)}$   &   $\textbf{0.15} (0.00)$   &   $\textbf{0.29 (0.01)}$   &   $0.55 (0.00)$   \\
FGP300   &   $0.43 (0.04)$   &   $0.38 (0.01)$   &   $0.42 (0.01)$   &   $0.73 (0.01)$   &   $0.17 (0.00)$   &   $0.36 (0.00)$   &   $0.55 (0.00)$   \\
MGP300   &   $\textbf{0.24 (0.00)}$   &   $\textbf{0.37 (0.01)}$   &   $\textbf{0.38 (0.01)}$   &   $0.70 (0.01)$   &   $0.16 (0.00)$   &   $0.33 (0.00)$   &   $\textbf{0.53 (0.00)}$   \\
\bottomrule
\end{tabular}
\end{sc}
\end{small}
\endgroup
\end{center}
\caption{RMSE (standard deviations reported in parentheses) on seven standard benchmarkreal-world datasets The lowest RMSE is in bold. The experimental set-ups are the same as in Table \ref{table:nlpd_time}.}
\label{table:rmse}
\end{table*}

\subsection{Curve learning via low-rank kernel approximations}\label{sub_sec:sim_data}
We examine the flexibility of our models by comparing them to exact Gaussian process regression models via the following simple example. We generate an artificial dataset based on the function $f(x) = \frac{1}{2} \left( 3 \sin(2 x) + \cos(10 x) + \frac{x}{4} \right)$; exact Gaussian process models can easily recover such a smooth function and, therefore, they provide a sound baseline for comparison with our methods. Our simulated dataset has one-dimensional training points $\{x_i, f(x_i)\}_{i=1}^{25}$ where $x_i \sim \mathcal{N}(0,1)$. For all three methods, a Gaussian kernel is used. The exact Gaussian process model has been trained using GPflow.

\begin{figure*}[h]
\centering
\includegraphics[width=14.2cm, height=15cm]{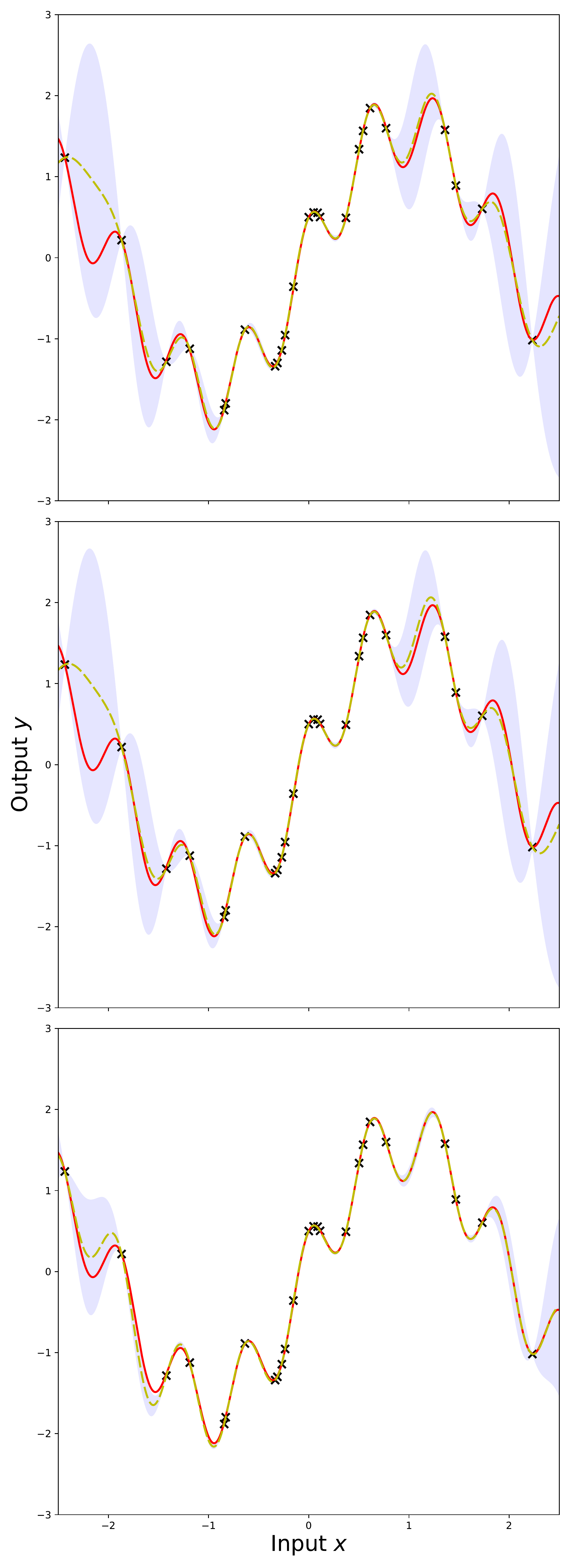}
\caption{Recovering the function $f(x) = \frac{1}{2} \left( 3 \sin(2 x) + \cos(10 x) + \frac{x}{4} \right)$. From top to bottom: Predictive mean and 95\% of the predictive probability mass of exact Gaussian process, MGP and FGP, respectively. We make use of 34 eigenfunctions for MGP and 34 spectral frequencies for FGP, i.e. $r=34$ and $r=68$, respectively. Black crosses depict the training data, solid red line shows $f(x)$, and the dashed yellow line shows the mean prediction of exact GP, MGP and FGP respectively.} 
\label{fig:1d_example_exact}
\end{figure*}

Figure \ref{fig:1d_example_exact} illustrates  how MGP and FGP compare to exact Gaussian process. MGP presents similar behavior, leading to similar posterior mean and predictive intervals. The posterior mean of FGP approximates better the underlying curve with more `confidence' to unseen function values. We use $r=34$ and $r=68$ for MGP and FGP respectively. 

\begin{figure*}[h]
\centering
\begin{tabular}{c}
{\includegraphics[width=0.9\textwidth, height=4.0cm]{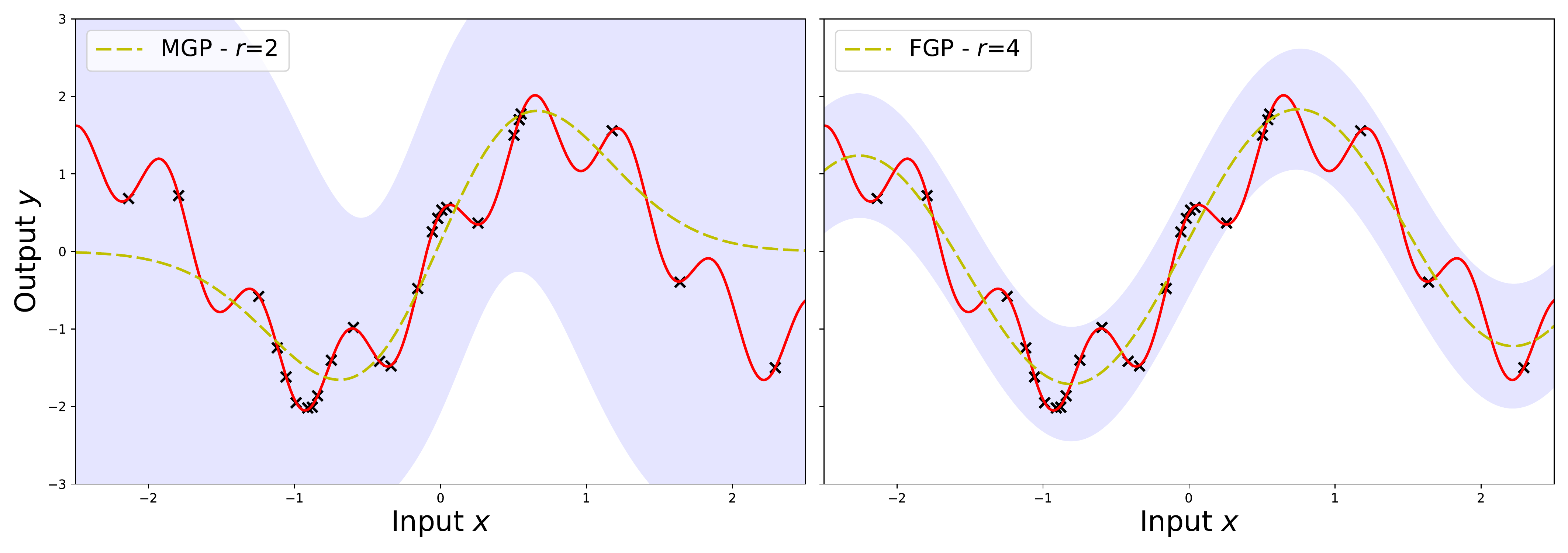}}  \\
{\includegraphics[width=0.9\textwidth, height=4.0cm]{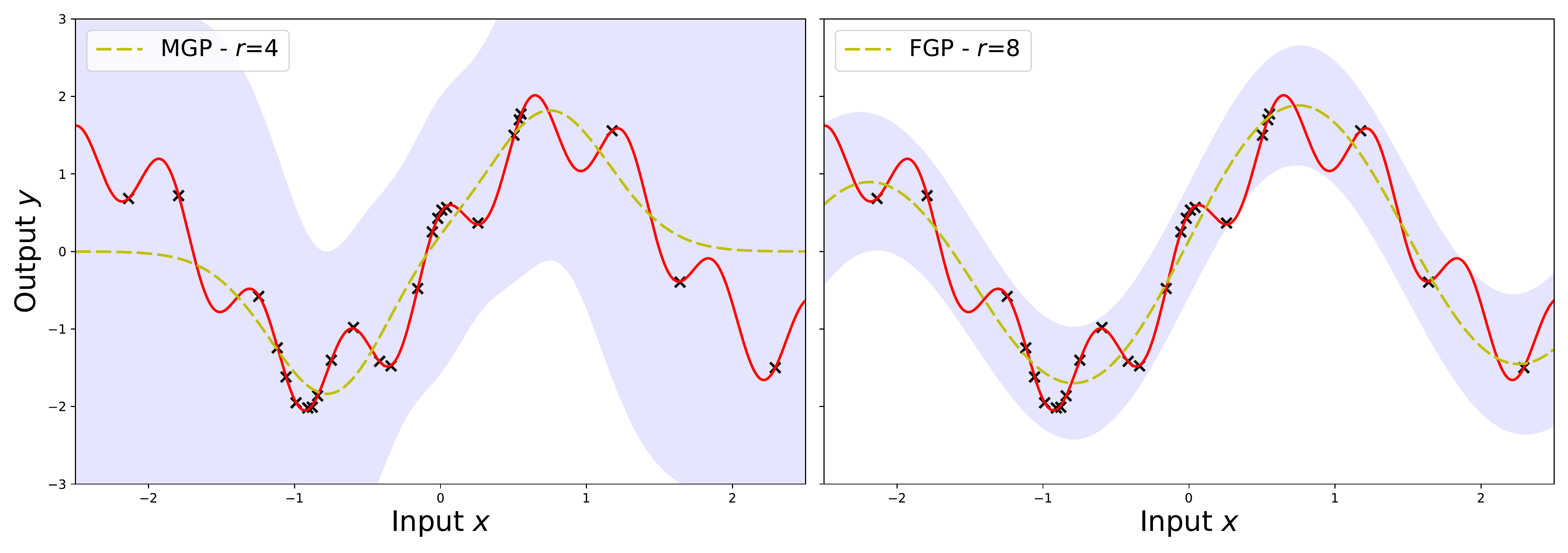}} \\
{\includegraphics[width=0.9\textwidth, height=4.0cm]{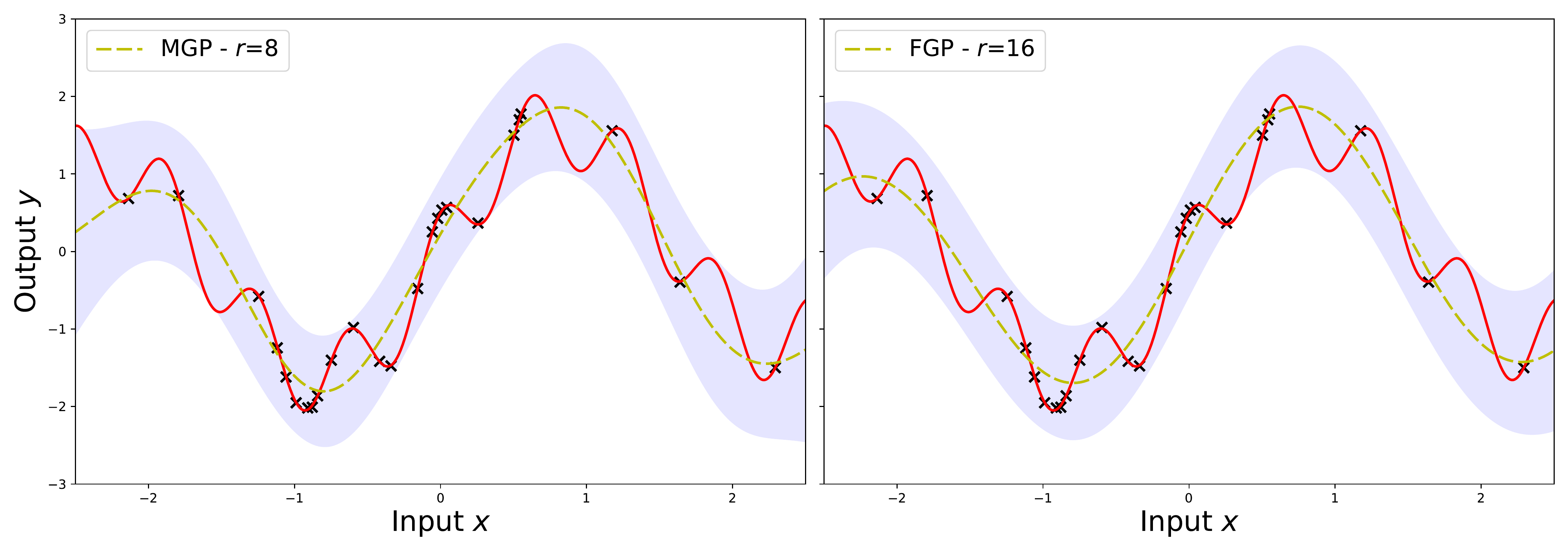}} \\
{\includegraphics[width=0.9\textwidth, height=4.0cm]{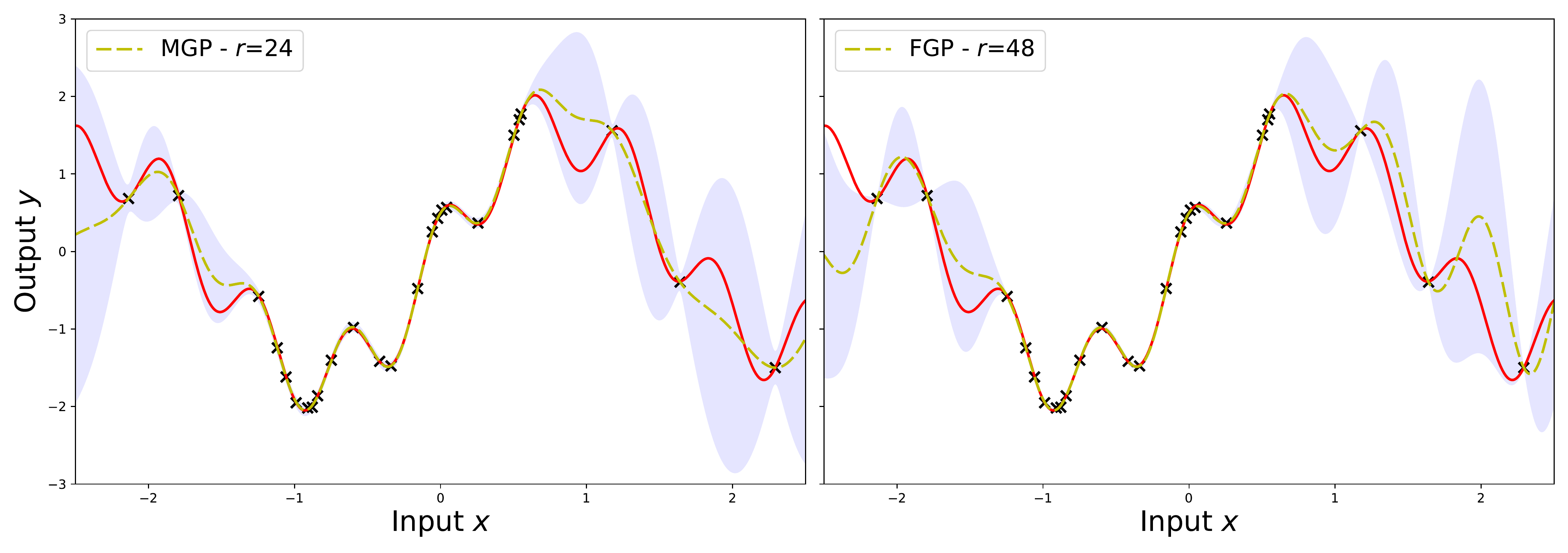}}  \\
{\includegraphics[width=0.9\textwidth, height=4.0cm]{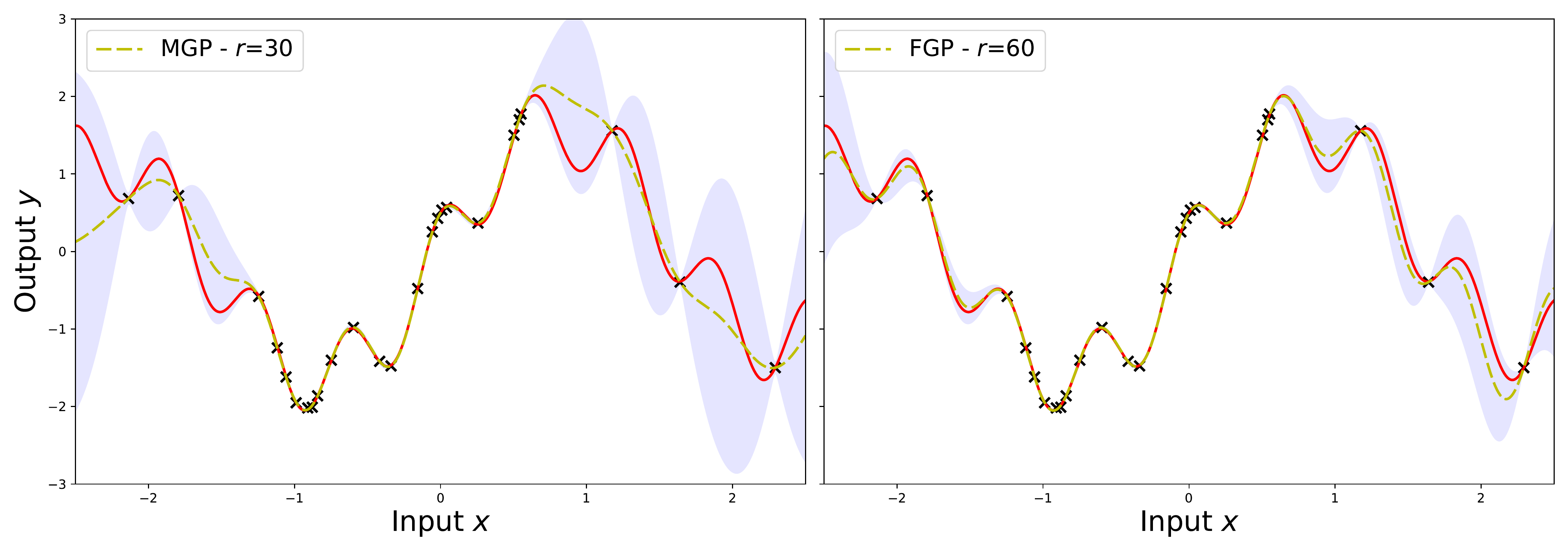}} \\
\end{tabular}
\caption{Recovering the function $f(x) = \frac{1}{2} \left( 3 \sin(2 x) + \cos(10 x) + \frac{x}{4} \right)$ when using different rank values $r$ to approximate the true kernel. Training points are denoted by black crosses,  $f(x)$ by solid red lines, MGP and FGP mean predictions with dashed yellow lines, and $95\%$ intervals of the predictive probability by shaded purple.}
\label{fig:1d_example_m}
\end{figure*}

Figure \ref{fig:1d_example_m} depicts how MGP and FGP inference is affected by considering different values for $r$ (i.e. eigenfunctions or  spectral frequencies) for approximating the kernel.
For MGP, as $r$ increases, the uncertainty decreases and posterior mean  estimates tend to approximate very well 
those of the exact GP model. FGP performs well with high confidence even with $r=4$ and after $r=24$ learns the true function accurately. 

\section{DNNs for feature extraction } \label{sec:idealized process}

We describe here an implementation architecture that exploits our theoretical guarantees with the enhancement of feature extraction through a DNN.
Instead of defining a direct mapping from $\bx \in \mathbb{R}^D$ to $y$ through a GP, we define a composition of a random function with a deterministic function as follows. 
First, a deterministic function $g_w: \bx \mapsto \bz$ embeds a feature vector $\bx$ to a feature vector $\bz \in \mathbb{R}^d$; we assume that $g_w$ is parametric, e.g.~expressible by a DNN.
Next, a random function $h:\bz \mapsto y$ is sampled from a GP with noisy observations exactly as described in Section~\ref{sec:approximation guarantees}, so $f(\cdot)$ is sampled from a GP with mean zero and kernel function $k_\theta:\mathbb{R}^d \times \mathbb{R}^d  \rightarrow \mathbb{R}$, and then $y \sim {\cal N}(f(\bz),\sigma^2)$, so 
\begin{align}
\by \sim {\cal N}(0,K(k_\theta,Z)+\sigma^2 I_N), \label{eq:idealized model-2}
\end{align}
where $Z =(g_w(\bx_i) \equiv \bz_i)_{i=1}^N$.
Clearly, by taking the neural network to be trivial (i.e.~the identity function) we obtain the setting of the Section \ref{sec:approximation guarantees}. 
The goal now is to identify a feature map $\phi_{\theta,\varepsilon}:\mathbb{R}^d \rightarrow \mathbb{R}^r$, providing a guarantee of the form 
\begin{align}
    K(k_\theta,Z) \approx_{\varepsilon} \Sigma(\phi_{\theta,\varepsilon},Z), \label{eq:low-rank kernel approximation guarantee-2}
\end{align}
where $\Sigma(\phi_{\theta,\varepsilon}, Z)=(\phi_{\theta,\varepsilon}(\bz_i)^{\top} \phi_{\theta,\varepsilon}(\bz_j))_{ij}$. 
With this DNN enhancement, using Random Fourier Features~\citep{rahimi2008random}, Modified Random Fourier Features~\citep{avron2017random}, or other random feature-based methods to obtain a low-rank approximation to the kernel $K(k_\theta,Z)$ gives rise to our family of {\em Deep Fourier Gaussian Processes (DFGP)}.  Using Mercer approximations gives rise to our family of {\em Deep Mercer Gaussian Processes (DMGP)}.

\subsection{Implementation details for DMGP and DFGP}\label{sec:implementations}

We provide implementation details on how we implement DMGP and DFGP using a Gaussian kernel. In both cases, the crux is to compute the low-rank matrix $\Sigma$ for a fixed rank $r$. For DMGP, we compute $\Sigma$ by using $\sqrt[d]{r} \in \mathbb{N}$ eigenfunctions/eigenvalues per dimension for the Mercer expansion in \eqref{kernel}:
$$\Sigma = \sum_{\bn \in \mathbb{N}^d, \bn \le (\sqrt[d]{r}, \ldots, \sqrt[d]{r})} \lambda_{\bn} \xi_{\bn} \xi_{\bn}^{\top}, $$ 
where $\xi_{\bn}= [ \phii_{\bn}(\bz_1), \ldots, \phii_{\bn}(\bz_N) ]^{\top} \in \mathbb{R}^N $ . Note that the parameter $a_j$ in \eqref{input} has to be pre-fixed or learnt from the data. We choose to keep it fixed with its value being set $1 / \sqrt{2}$ which corresponds to a standard $d$-dimensional Gaussian measure and we standardize the outputs of DNN, $Z$, before we feed it as an input to the GP.

Regarding DFGP, we follow the implementation based in algorithm 1 of \cite{rahimi2008random}, where we first sample, for even number $r$, $r \over 2$ spectral frequencies $\boldeta_1,\ldots,\boldeta_{r \over 2}$ from the spectral density $p(\boldeta)$ of the 
stationary kernel $k_{\theta} (\cdot , \cdot)$ and then create the feature map $\phi(\bz):\mathbb{R}^d \rightarrow \mathbb{R}^{r}$, defined by
the vector $$\sqrt{\frac{2}{r}}[\cos(\boldeta_1^{\top} \bz),\ldots, \cos(\boldeta_{r \over 2}^{\top} \bz), \sin(\boldeta_1^{\top} \bz),\ldots, \sin(\boldeta_{r \over 2}^{\top} \bz) ]^{\top}.$$ 
Hence, the rank of $\Sigma$ is always an even number. The spectral frequencies are only sampled once before training and are then kept fixed throughout optimization of the log-marginal likelihood. Finally, the spectral density in the case of Gaussian kernel in \eqref{eq:ard_kernel} is given by
$ p(\boldeta) =  \sqrt{| 2 \pi \Delta^{-1} |}\sigma_f^{-2}  \exp( - 2 \pi^2 \boldeta^{\top} \Delta^{-1} \boldeta )$.

\begin{figure*}
\centering
\includegraphics[width=14cm, height=4.2cm]{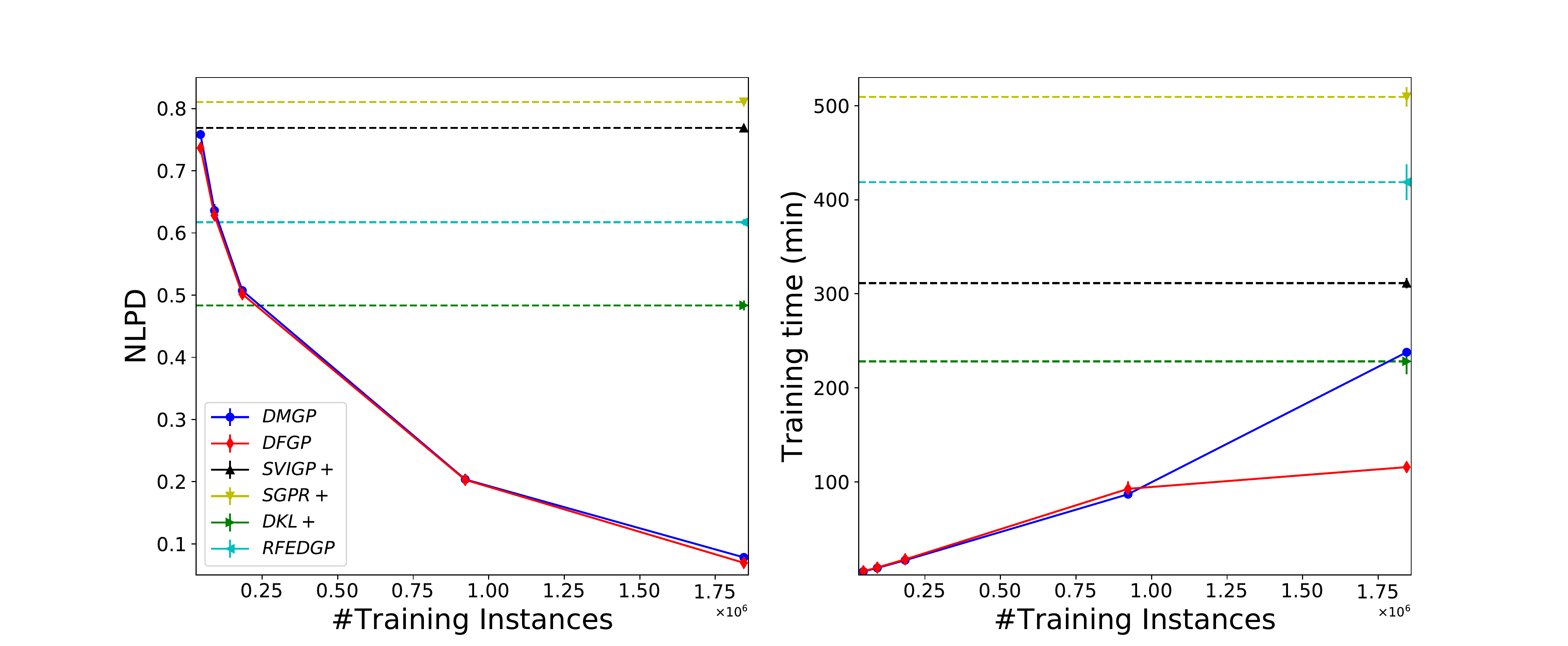} 
\caption{Negative log-predictive density (left) and training times (right)  as a function of the number of training points  for the {\sc Electric} dataset. Dashed lines correspond to baseline models trained on the full dataset and their values can be also found in Table \ref{table:nlpd_time}. } 
\label{fig:electric_ratio}
\end{figure*}

\subsection{Experimental evaluation}\label{sec:experiments_dmgp_dfgp}
We compare the following methods: (i) DMGP with $d=1$ and $r=15$; (ii) DFGP with 
$d=4$, $r=40$, and random Fourier features; (iii) Stochastic Variational Inference GP with $250$ (\emph{SVIGP}) and $500$ (\emph{SVIGP+}) inducing points \cite{hensman2013gaussian} (code used from GPflow \citep{GPflow2017}); (iv) Sparse GP Regression  \citep{titsias2009variational} with $250$ (\emph{SGPR}) and $500$ (\emph{SGPR+}) inducing points  (code used from GPflow);
(v) Deep Kernel Learning with $5000$ (\emph{DKL}) and $10000$ (\emph{DKL+}) inducing points  and $d=1$ since we found that larger values of $d$ did not improve performance (code used from \url{https://gpytorch.ai}) \cite{wilson2016deep}; (vi) Deep GPs with random Fourier features (\emph{RFEDGP}), see \cite{cutajar2017random}, with two hidden layers, three GPs per layer, and spectral frequencies being optimized variationally with fixed randomness; we used $20$ Monte Carlo samples throughout training since we found it is much faster and as accurate as the training procedure followed by \cite{cutajar2017random} and $100$ Monte Carlo samples for prediction as in \cite{cutajar2017random} (code used from \url{https://github.com/mauriziofilippone/deep_gp_random_features}). All data  have been retrieved from UCI repository \citep{Dua2019UCI} or the official site of \cite{williams2006gaussian}.

DMGP and DFGP require joint estimation of the parameters $w$ and $\theta$ through maximization of the  log marginal likelihood which is a non-decomposable loss function, see \cite{kar2014online}, so we used the semi-stochastic asynchronous gradient descent suggested in \cite{al2017learning}. More details about the practical implementation of DMGP and DFGP are discussed in Supplement  \ref{sec:implementations}. We emphasise that for maintaining fairness among comparisons, we kept hyperparameter tuning to the minimum for the DNN-based methods, by using, across all datasets, the same [$D$ \textendash ~512 \textendash ~256 \textendash ~64 \textendash ~$d$  ] architecture with hyperbolic tangent activation functions, while the DNN weights of these methods were initialized by pre-training the DNN as suggested by  \cite{wilson2016deep,wilson2016stochastic}.  
We ran all methods for $100$ epochs using Adam optimizer \citep{kingma2014adam} and mini-batch optimization with mini-batches of size $1000$. All GPs used Gaussian kernels with separate length-scale per dimension. All results have been averaged over five random splits (90\% train, 10\% test).

\begin{table*}[h!]
\begin{center}
\begingroup
\setlength{\tabcolsep}{2.1pt} 
\renewcommand{\arraystretch}{1}
\begin{tiny}
\begin{sc}
\begin{tabular}{l c c c c c c c}
\toprule
&  \multicolumn{6}{c}{\textbf{Negative Log-Predictive Density}}  \\
\cmidrule(lr){2-8}
   &   Elevators   &   Protein   &   Sarcos   &   3DRoad   &   Song   &   Buzz   &   Electric   \\
$N$ &   $14939 $   &   $41157$   &   $ 44039 $   &   $ 391386 $   &   $463810 $   &   $524925 $   &   $1844352 $   \\
 $N^*$ &   $1660 $   &   $4573 $   &   $ 4894 $   &   $ 43488 $   &   $51535 $   &   $58325 $   &   $204928 $   \\
 $D$ &   $ 18 $   &   $9 $   &   $ 21 $   &   $ 3 $   &   $90 $   &   $77 $   &   $19 $   \\
\midrule
SVIGP   &   $0.444 (0.021)$   &   $1.041 (0.007)$   &   $-0.422 (0.006)$   &   $0.652 (0.008)$   &   $1.208 (0.005)$   &   $0.087 (0.006)$   &   $0.804 (0.003)$   \\
SVIGP+   &   $0.435 (0.018)$   &   $0.991 (0.006)$   &   $-0.479 (0.004)$   &   $0.541 (0.008)$   &   $1.205 (0.005)$   &   $0.078 (0.005)$   &   $0.769 (0.002)$   \\
SGPR   &   $0.433 (0.017)$   &   $0.997 (0.007)$   &   $-0.370 (0.007)$   &   $0.799 (0.007)$   &   $1.202 (0.006)$   &   $0.216 (0.005)$   &   $0.871 (0.002)$   \\
SGPR+   &   $0.420 (0.017)$   &   $0.944 (0.005)$   &   $-0.468 (0.009)$   &   $0.737 (0.011)$   &   $1.198 (0.006)$   &   $0.186 (0.004)$   &   $0.810 (0.001)$   \\
DKL   &   $0.527 (0.011)$   &   $0.958 (0.020)$   &   $0.395 (0.040)$   &   $0.744 (0.129)$   &   $1.261 (0.057)$   &   $0.460 (0.003)$   &   $0.447 (0.013)$   \\
DKL+   &   $0.536 (0.011)$   &   $0.961 (0.037)$   &   $0.430 (0.034)$   &   $0.687 (0.047)$   &   $1.315 (0.158)$   &   $0.438 (0.017)$   &   $0.448 (0.012)$   \\
RFEDGP   &   $0.434 (0.021)$   &   $1.028 (0.006)$   &   $-0.303 (0.061)$   &   $0.583 (0.009)$   &   $1.207 (0.006)$   &   $0.238 (0.032)$   &   $0.616 (0.004)$   \\
DMGP   &   $0.371 (0.036)$   &   $0.857 (0.015)$   &   $\mathbf{-0.777} (0.015)$   &   $0.140 (0.010)$   &   $\mathbf{1.185} (0.004)$   &   $-0.008 (0.022)$   &   $0.078 (0.002)$   \\
DFGP   &   $\mathbf{0.350} (0.029)$   &   $\mathbf{0.853} (0.018)$   &   $\mathbf{-0.777} (0.020)$   &   $\mathbf{0.139} (0.012)$   &   $1.189 (0.005)$   &   $\mathbf{-0.016} (0.002)$   &   $\mathbf{0.067} (0.004)$   \\
\midrule
DNN+S   &   $0.402 (0.030)$   &   $0.904 (0.013)$   &   $-0.559 (0.021)$   &   $0.239 (0.020)$   &   $1.211 (0.001)$   &   $0.019 (0.003)$   &   $0.165 (0.001)$   \\
DNN+M   &   $0.401 (0.030)$   &   $0.893 (0.016)$   &   $-0.585 (0.029)$   &   $0.233 (0.020)$   &   $1.208 (0.001)$   &   $0.025 (0.016)$   &   $0.164 (0.001)$   \\
DNN+F   &   $0.380 (0.022)$   &   $0.895 (0.022)$   &   $-0.628 (0.044)$   &   $0.237 (0.008)$   &   $1.210 (0.002)$   &   $0.012 (0.001)$   &   $0.155 (0.001)$   \\
\midrule
\midrule
&  \multicolumn{6}{c}{\textbf{Training Time (seconds)}}  \\
\midrule
SVIGP   &   $59 (2)$   &   $182 (24)$   &   $269 (19)$   &   $2096 (297)$   &   $2527 (19)$   &   $2615 (165)$   &   $8231 (302)$   \\
SVIGP+   &   $150 (5)$   &   $425 (3)$   &   $455 (2)$   &   $3895 (92)$   &   $4845 (132)$   &   $5715 (102)$   &   $18878 (1513)$   \\
SGPR   &   $\mathbf{49} (1)$   &   $156 (15)$   &   $227 (11)$   &   $1697 (53)$   &   $2012 (13)$   &   $2114 (109)$   &   $11190 (68)$   \\
SGPR+   &   $144 (3)$   &   $381 (11)$   &   $419 (7)$   &   $3661 (124)$   &   $4676 (118)$   &   $5208 (336)$   &   $30357 (573)$   \\
DKL   &   $285 (27)$   &   $435 (4)$   &   $455 (5)$   &   $2531 (31)$   &   $2916 (180)$   &   $2854 (408)$   &   $14455 (596)$   \\
DKL+   &   $774 (80)$   &   $1317 (227)$   &   $740 (402)$   &   $2377 (194)$   &   $2885 (161)$   &   $3182 (245)$   &   $14833 (1608)$   \\
RFEDGP   &   $184 (9)$   &   $559 (43)$   &   $629 (49)$   &   $2862 (296)$   &   $4627 (66)$   &   $4276 (232)$   &   $26256 (1647)$   \\
DMGP   &   $121 (26)$   &   $375 (23)$   &   $448 (26)$   &   $3602 (238)$   &   $3598 (95)$   &   $3963 (63)$   &   $14311 (134)$   \\
DFGP   &   $51 (2)$   &   $\mathbf{137} (13)$   &   $\mathbf{146} (1)$   &   $\mathbf{1363} (8)$   &   $\mathbf{1898} (15)$   &   $\mathbf{2092} (26)$   &   $\mathbf{6785} (323)$   \\
\midrule
DNN+S   &   $28 (2)$   &   $80 (7)$   &   $78 (4)$   &   $752 (57)$   &   $224 (8)$   &   $513 (7)$   &   $2211 (614)$   \\
DNN+M   &   $41 (4)$   &   $113 (10)$   &   $113 (7)$   &   $1061 (79)$   &   $331 (13)$   &   $711 (12)$   &   $3069 (841)$   \\
DNN+F   &   $53 (3)$   &   $150 (17)$   &   $171 (13)$   &   $1326 (93)$   &   $245 (13)$   &   $1890 (133)$   &   $6504 (294)$   \\
\bottomrule
\end{tabular}
\end{sc}
\end{tiny}
\endgroup
\end{center}
\caption{Negative log-predictive density and training time comparison (standard deviations reported in parentheses) on seven standard benchmark real-world datasets; $N,N^*$ and $D$ represent training data size, test data size, and feature dimension, respectively.} 
\label{table:nlpd_time_dnn}
\end{table*}

Table \ref{table:nlpd_time_dnn} presents comparisons of all methods in terms of NLPD and training time, whereas Table \ref{table:rmse_dnn} presents comparisons in terms of RMSE, which carry the same message.  Both DFGP and DMGP clearly outperform all other methods in speed and NLPD performance. The last three rows of the two sub-tables of Table \ref{table:nlpd_time} describe results of extra experiments in which a DNN regression model with RMSE as loss function was first trained on the data, then its fitted outputs $Z$ were independently used as input to fit a Mercer GP (\emph{DNN+M}), random Fourier features GP (\emph{DNN+F}), or simply an isotropic model $\by \sim \mathcal{N}(Z,\sigma^2 I_N )$ (\emph{DNN+S}). These methods do not perform as well in terms of NLPD, emphasizing the necessity of our suggested joint parameter optimization.  However, notice the improvement of the non-parametric \emph{DNN+M} and \emph{DNN+F} over the naive \emph{DNN+S}.  We also applied an exact GP regression model (using GPflow)  to the  smallest dataset {\sc Elevators}. The average NLPD ($\pm$ one st.d.) was $0.377 \pm 0.024$ with total average running time $53550 \pm 2099$ seconds. Comparing with the results of Table \ref{table:nlpd_time_dnn} we see that both DFGP and NLPD exhibited superior NLPD performance confirming the effectiveness of DNN feature engineering.

Figure \ref{fig:electric_ratio} depicts how NLPD and training time over 100 epochs depend on the number of training points in the {\sc Electric} dataset, illustrating that our methods can achieve equally good precision with less training points and less time. In particular, notice that DFGP scales better than DMGP.  Table \ref{table:results_nlpd_time_dm} presents the performance of DMGP and DFGP for a series of values of $d$ and $r$, for the smaller  size datasets {\sc Protein} and {\sc Sarcos}.  Similar results for {\sc Elevators} dataset can be found in Table \ref{table:results_nlpd_time_dm_elevators}.  There is evidence that large values of $d$ and $r$ offer only marginally better performance for both DMGP and DFGP, while severely affecting the training 
time for DMGP.  This suggests using relatively small $d$ and $r$ for DMGP  and slightly increase these values for DFGP. For all our data experiments we used $d=1, r=15$ for DMGP and $d=4, r=40$ for DFGP.

\begin{table*}
\begin{center}
\begingroup
\setlength{\tabcolsep}{2.8pt} 
\renewcommand{\arraystretch}{1}
\begin{tiny}
\begin{sc}
\begin{tabular}{lccccccc}
\toprule
&  \multicolumn{6}{c}{\textbf{RMSE}}  \\
\cmidrule(lr){2-8}
   &   Elevators   &   Protein   &   Sarcos   &   3DRoad   &   Song   &   Buzz   &   Electric   \\
$N$ &   $14939 $   &   $41157$   &   $ 44039 $   &   $ 391386 $   &   $463810 $   &   $524925 $   &   $1844352 $   \\
 $N^*$ &   $1660 $   &   $4573 $   &   $ 4894 $   &   $ 43488 $   &   $51535 $   &   $58325 $   &   $204928 $   \\
 $D$ &   $ 18 $   &   $9 $   &   $ 21 $   &   $ 3 $   &   $90 $   &   $77 $   &   $19 $   \\
\midrule
SVIGP   &   $0.379 (0.009)$   &   $0.683 (0.005)$   &   $0.160 (0.001)$   &   $0.462 (0.004)$   &   $0.810 (0.005)$   &   $0.271 (0.003)$   &   $0.540 (0.002)$   \\
SVIGP+   &   $0.375 (0.007)$   &   $0.649 (0.005)$   &   $0.151 (0.001)$   &   $0.413 (0.004)$   &   $0.807 (0.004)$   &   $0.270 (0.003)$   &   $0.521 (0.001)$   \\
SGPR   &   $0.375 (0.007)$   &   $0.653 (0.005)$   &   $0.168 (0.002)$   &   $0.537 (0.004)$   &   $0.806 (0.005)$   &   $0.315 (0.003)$   &   $0.577 (0.001)$   \\
SGPR+   &   $0.370 (0.007)$   &   $0.620 (0.004)$   &   $0.153 (0.002)$   &   $0.506 (0.006)$   &   $0.802 (0.005)$   &   $0.308 (0.003)$   &   $0.542 (0.001)$   \\
DKL   &   $0.352 (0.010)$   &   $0.630 (0.012)$   &   $0.230 (0.047)$   &   $0.499 (0.074)$   &   $0.815 (0.006)$   &   $0.274 (0.014)$   &   $0.285 (0.008)$   \\
DKL+   &   $0.361 (0.009)$   &   $0.632 (0.022)$   &   $0.276 (0.035)$   &   $0.474 (0.024)$   &   $0.813 (0.004)$   &   $0.268 (0.014)$   &   $0.296 (0.015)$   \\
RFEDGP   &   $0.355 (0.013)$   &   $0.678 (0.004)$   &   $0.179 (0.012)$   &   $0.434 (0.004)$   &   $0.809 (0.005)$   &   $0.307 (0.009)$   &   $0.448 (0.002)$   \\
DMGP   &   $0.346 (0.010)$   &   $0.564 (0.007)$   &   $\mathbf{0.111} (0.002)$   &   $\mathbf{0.277} (0.003)$   &   $\mathbf{0.791} (0.003)$   &   $\mathbf{0.237} (0.000)$   &   $0.261 (0.001)$   \\
DFGP   &   $\mathbf{0.341} (0.008)$   &   $\mathbf{0.562} (0.008)$   &   $\mathbf{0.111} (0.002)$   &   $0.278 (0.003)$   &   $0.795 (0.004)$   &   $0.238 (0.000)$   &   $\mathbf{0.259} (0.001)$   \\
\midrule
DNN+S   &   $0.359 (0.007)$   &   $0.588 (0.006)$   &   $0.144 (0.001)$   &   $0.311 (0.005)$   &   $0.806 (0.001)$   &   $0.251 (0.000)$   &   $0.288 (0.001)$   \\
DNN+M   &   $0.359 (0.007)$   &   $0.581 (0.006)$   &   $0.140 (0.003)$   &   $0.310 (0.005)$   &   $0.804 (0.001)$   &   $0.250 (0.001)$   &   $0.287 (0.001)$   \\
DNN+F   &   $0.354 (0.005)$   &   $0.582 (0.009)$   &   $0.135 (0.004)$   &   $0.311 (0.001)$   &   $0.805 (0.002)$   &   $0.250 (0.001)$   &   $0.285 (0.001)$   \\
\bottomrule
\end{tabular}
\end{sc}
\end{tiny}
\endgroup
\end{center}
\caption{RMSE comparison between state-of-the-art baselines and our methods DMGP and DFGP. The experimental set-ups are the same as in Table \ref{table:nlpd_time_dnn} of the main paper.}
\label{table:rmse_dnn}
\end{table*}

\subsection{Summary of results}
The extensive experiments of this section were designed to answer specific performance questions, the answers to which are summarized here. There is strong evidence that both instantiations of our  framework, DFGP and DMGP, (i) outperform all state-of-the-art baselines in both time efficiency and prediction accuracy measured in NLPD and RMSE, (ii) outperform simple DNN regression without the use of a GP verifying the need for incorporating both our proposed  ingredients, (iii) achieve competitive  performance and are much faster against the competitors with quite fewer training points, (iv) outperform exact GP regression inference confirming the importance of the DNN feature extraction, and (v) illustrate the importance of our proposed joint parameter estimation  framework since they clearly outperform consecutive estimation of the DNN first and the kernel parameters after. We also illustrate robustness with respect to $r$ and $d$ and provide  practical guidelines.

\begin{table*}[h!]
\begin{center}
\begingroup
\setlength{\tabcolsep}{6.3pt} 
\renewcommand{\arraystretch}{1}
\begin{tiny}
\begin{sc}
\begin{tabular}{lcccccc}
\toprule
 &  \multicolumn{3}{c}{{\sc Protein}} & \multicolumn{3}{c}{{\sc Sarcos}} \\
\cmidrule(lr){2-4}\cmidrule(lr){5-7}
 &  \multicolumn{6}{c}{\textbf{Negative log-predictive density}-\textbf{DMGP}}  \\
\cmidrule(lr){2-7}
$\sqrt[d]{r}$~~~  & \multicolumn{1}{c}{$d=1$} & \multicolumn{1}{c}{$d=2$} & \multicolumn{1}{c}{$d=3$} & \multicolumn{1}{c}{$d=1$} & \multicolumn{1}{c}{$d=2$} & \multicolumn{1}{c}{$d=3$} \\
\midrule
$2$   & $0.883 (0.014)$   & $0.872 (0.012)$   & $0.872 (0.015)$   & $-0.778 (0.012)$   & $-0.762 (0.019)$   & $-0.754 (0.029)$    \\
$4$   & $0.856 (0.015)$   & $0.863 (0.014)$   & $0.867 (0.025)$   & $-0.777 (0.015)$   & $-0.775 (0.019)$   & $-0.780 (0.016)$    \\
$8$   & $0.857 (0.015)$   & $0.855 (0.013)$   & $0.848 (0.014)$   & $-0.778 (0.015)$   & $-0.773 (0.021)$   & $-0.780 (0.019)$    \\
$10$   & $0.857 (0.015)$   & $0.855 (0.013)$   & $0.848 (0.014)$   & $-0.777 (0.015)$   & $-0.772 (0.021)$   & $-0.780 (0.020)$    \\
$16$   & $0.857 (0.015)$   & $0.855 (0.013)$   & $0.848 (0.015)$   & $-0.778 (0.015)$   & $-0.772 (0.021)$   & $-0.770 (0.020)$    \\
$32$   & $0.857 (0.015)$   & $0.855 (0.013)$   & \textendash   & $-0.777 (0.015)$   & $-0.772 (0.021)$   & \textendash    \\
\midrule
$\frac{r}{2}$ &  \multicolumn{6}{c}{\textbf{Negative log-predictive density}-\textbf{DFGP}}  \\
\midrule
$2$   & $0.871 (0.013)$   & $0.873 (0.014)$   & $0.862 (0.013)$   & $-0.608 (0.130)$   & $-0.697 (0.069)$   & $-0.771 (0.019)$    \\
$4$   & $0.856 (0.013)$   & $0.851 (0.012)$   & $0.847 (0.014)$   & $-0.784 (0.014)$   & $-0.778 (0.021)$   & $-0.783 (0.019)$    \\
$8$   & $0.856 (0.014)$   & $0.854 (0.012)$   & $0.846 (0.013)$   & $-0.784 (0.014)$   & $-0.779 (0.021)$   & $-0.784 (0.020)$    \\
$10$   & $0.856 (0.014)$   & $0.855 (0.013)$   & $0.846 (0.014)$   & $-0.784 (0.014)$   & $-0.779 (0.020)$   & $-0.786 (0.020)$    \\
$16$   & $0.856 (0.014)$   & $0.854 (0.012)$   & $0.846 (0.015)$   & $-0.784 (0.014)$   & $-0.779 (0.021)$   & $-0.784 (0.022)$    \\
$32$   & $0.856 (0.014)$   & $0.853 (0.012)$   & $0.847 (0.015)$   & $-0.785 (0.014)$   & $-0.781 (0.021)$   & $-0.785 (0.019)$    \\
\midrule
\midrule
$\sqrt[d]{r}$ &  \multicolumn{6}{c}{\textbf{Training Time}-\textbf{DMGP}}  \\
\midrule
$2$   & $115 (4)$   & $132 (2)$   & $154 (2)$   & $127 (3)$   & $144 (3)$   & $174 (3)$    \\
$4$   & $112 (1)$   & $170 (3)$   & $374 (12)$   & $127 (6)$   & $187 (2)$   & $417 (21)$    \\
$8$   & $114 (1)$   & $308 (9)$   & $874 (26)$   & $124 (6)$   & $325 (11)$   & $980 (18)$    \\
$10$   & $116 (5)$   & $369 (11)$   & $2147 (63)$   & $128 (5)$   & $401 (15)$   & $2325 (84)$    \\
$16$   & $117 (1)$   & $404 (12)$   & $88649 (163)$   & $130 (4)$   & $456 (16)$   & $94965 (293)$    \\
$32$   & $122 (1)$   & $1864 (21)$   & \textendash   & $135 (6)$   & $2071 (72)$   & \textendash    \\
\midrule
$\frac{r}{2} $ &  \multicolumn{6}{c}{\textbf{Training Time}-\textbf{DFGP}}  \\
\midrule
$2$   & $108 (1)$   & $108 (1)$   & $108 (1)$   & $124 (3)$   & $121 (3)$   & $124 (3)$    \\
$4$   & $109 (1)$   & $111 (2)$   & $110 (2)$   & $123 (7)$   & $130 (2)$   & $126 (4)$    \\
$8$   & $112 (1)$   & $112 (1)$   & $113 (2)$   & $125 (2)$   & $132 (1)$   & $133 (2)$    \\
$10$   & $115 (4)$   & $113 (6)$   & $126 (3)$   & $127 (4)$   & $134 (5)$   & $136 (3)$    \\
$16$   & $118 (1)$   & $118 (2)$   & $156 (17)$   & $129 (8)$   & $136 (4)$   & $188 (9)$    \\
$32$   & $126 (1)$   & $126 (2)$   & $176 (20)$   & $141 (5)$   & $145 (4)$   & $199 (5)$    \\
\bottomrule
\end{tabular}
\end{sc}
\end{tiny}
\endgroup
\end{center}
\caption{Comparative negative log-predictive density performance and training time in seconds for different values of rank $r$ and embedding dimension $d$; standard deviations in parentheses. No results are reported for DMGP for $d=3, \sqrt[3]{r} = 32$ since computational tractability breaks for these values.}
\label{table:results_nlpd_time_dm}
\end{table*}

\begin{table*}
\vskip 0.15in
\begin{center}
\begingroup
\setlength{\tabcolsep}{6.53pt} 
\renewcommand{\arraystretch}{1}
\begin{tiny}
\begin{sc}
\begin{tabular}{lccccccc}
\toprule
  \multicolumn{8}{c}{{\sc Elevators}} \\
\cmidrule(lr){1-8}
   \multicolumn{4}{c}{\textbf{DMGP}}  &  \multicolumn{4}{c}{\textbf{DFGP}}   \\
\cmidrule(lr){1-4}\cmidrule(lr){5-8}
$\sqrt[d]{r}$  & \multicolumn{1}{c}{$d=1$} & \multicolumn{1}{c}{$d=2$} & \multicolumn{1}{c}{$d=3$} & 
$r \over 2$ &
\multicolumn{1}{c}{$d=1$} & \multicolumn{1}{c}{$d=2$} & \multicolumn{1}{c}{$d=3$} \\
\midrule
 &  \multicolumn{6}{c}{\textbf{NLPD}} \\
\midrule
$2$   & $0.381 (0.037)$   & $0.361 (0.032)$   & $0.377 (0.044)$ & $2$  & $0.411 (0.037)$   & $0.381 (0.040)$   & $0.367 (0.029)$   \\
$4$   & $0.371 (0.036)$   & $0.351 (0.032)$   & $0.353 (0.028)$ & $4$  & $0.380 (0.037)$   & $0.357 (0.032)$   & $0.357 (0.030)$    \\
$8$   & $0.371 (0.036)$   & $0.351 (0.032)$   & $0.352 (0.029)$ & $8$  & $0.379 (0.036)$   & $0.356 (0.032)$   & $0.356 (0.030)$    \\
$10$   & $0.371 (0.037)$   & $0.351 (0.032)$   & $0.352 (0.029)$ & $10$  & $0.379 (0.036)$   & $0.357 (0.031)$   & $0.357 (0.029)$    \\
$16$   & $0.371 (0.036)$   & $0.351 (0.032)$   & $0.352 (0.029)$ & $16$  & $0.379 (0.037)$   & $0.357 (0.032)$   & $0.357 (0.030)$    \\
$32$   & $0.371 (0.036)$   & $0.351 (0.032)$   & \textendash & $32$   & $0.379 (0.036)$   & $0.356 (0.032)$   & $0.357 (0.030)$    \\
\midrule
\midrule
&  \multicolumn{6}{c}{\textbf{Training Time}}  \\
\midrule
$2$   & $40 (1)$   & $49 (1)$   & $59 (1)$ & $2$  & $39 (1)$   & $38 (1)$   & $40 (0)$    \\
$4$   & $41 (1)$   & $58 (1)$   & $135 (2)$   & $4$ & $39 (1)$   & $38 (0)$   & $40 (0)$  \\
$8$   & $41 (2)$   & $100 (1)$   & $303 (5)$  & $8$ & $40 (1)$   & $39 (1)$   & $41 (0)$    \\
$10$   & $41 (2)$   & $117 (2)$   & $710 (19)$ & $10$  & $41 (2)$   & $40 (1)$   & $42 (1)$    \\
$16$   & $42 (1)$   & $140 (3)$   & $31593 (151)$ & $16$  & $42 (2)$   & $41 (0)$   & $51 (3)$    \\
$32$   & $44 (2)$   & $620 (10)$   & \textendash & $32$   & $46 (2)$   & $44 (1)$   & $55 (3)$    \\
\bottomrule
\end{tabular}
\end{sc}
\end{tiny}
\endgroup
\end{center}
\caption{Comparative NLPD performance and training time (in seconds) of {\sc DMGP} and {\sc DFGP} on {\sc Elevators} dataset for several values of rank $r$. No results are reported for DMGP for $d=3, \sqrt[3]{r} = 32$ since computational tractability breaks for these values. Experimental set-ups are the same as in Table 1 of the main paper.}
\label{table:results_nlpd_time_dm_elevators}
\end{table*}

\section{Code}

All experiments were carried out on a Linux machine with 32 2.20GHz CPU cores and 64GB RAM. The implementation of our code is available at \url{https://github.com/aresPanos/gurantees_GPR}.

\end{document}